\pdfoutput=1

\documentclass[11pt]{article}

\usepackage[final]{acl}

\usepackage{times}
\usepackage{multirow}
\usepackage{latexsym}

\usepackage[T1]{fontenc}

\usepackage[utf8]{inputenc}

\usepackage{microtype}

\usepackage{inconsolata}

\usepackage{graphicx}
\usepackage{url}
\usepackage[T1]{fontenc}
\usepackage{float}
\usepackage{graphicx}
\usepackage{booktabs}
\usepackage{siunitx}
\usepackage{hhline}
\usepackage{subcaption}
\usepackage{amsmath}
\usepackage{xspace,mfirstuc,tabulary}
\title{FaithfulSAE: Towards Capturing Faithful Features with Sparse Autoencoders without External Dataset Dependencies}

\author{
  \textbf{Seonglae Cho},
  \textbf{Harryn Oh},
  \textbf{Donghyun Lee},
  \textbf{Luis Eduardo Rodrigues Vieira},\\
  \textbf{Andrew Bermingham},
  \textbf{Ziad El Sayed}\\
  University College London\thanks{\{seonglae.cho.24, harryn.oh.21, donghyun.lee.21, luis.vieira.21, andrew.bermingham.24, ziad.sayed.24\}@ucl.ac.uk}
}

\begin{document}
\maketitle
\begin{abstract}
Sparse Autoencoders (SAEs) have emerged as a promising solution for decomposing large language model representations into interpretable features.
However, \citet{paulo2025sparseautoencoderstraineddata} have highlighted instability across different initialization seeds, and \citet{heap2025sparseautoencodersinterpretrandomly} have pointed out that SAEs may not capture model-internal features.
These problems likely stem from training SAEs on external datasets---either collected from the Web or generated by another model---which may contain out-of-distribution (OOD) data beyond the model's generalisation capabilities.
This can result in hallucinated SAE features, which we term "Fake Features", that misrepresent the model's internal activations.
To address these issues, we propose FaithfulSAE, a method that trains SAEs on the model's own synthetic dataset.
Using FaithfulSAEs, we demonstrate that training SAEs on less-OOD instruction datasets results in SAEs being more stable across seeds.
Notably, FaithfulSAEs outperform SAEs trained on web-based datasets in the SAE probing task and exhibit a lower Fake Feature Ratio in 5 out of 7 models.
Overall, our approach eliminates the dependency on external datasets, advancing interpretability by better capturing model-internal features while highlighting the often neglected importance of SAE training datasets.
\end{abstract}

\section{Introduction}
Sparse Autoencoders (SAEs), an architecture introduced by \citealp{faruqui-etal-2015-sparse}, have demonstrated the ability to transform Large Language Model (LLM) representations into interpretable features without supervision \citep{huben2024sparse}.
SAE latent dimensions can be trained to reconstruct activations while incurring a sparsity penalty, ideally resulting in a sparse mapping of human-interpretable features.
This approach enables decomposition of latent representations into interpretable features by reconstructing transformer hidden states \citep{gao2025scaling} or MLP activations \citep{bricken2023monosemanticity}.

Despite the demonstrated utility of SAE features, several concerns persist: SAEs can yield very different feature sets depending on the initialization seed \citep{paulo2025sparseautoencoderstraineddata}, SAEs can exhibit highly activated latents which reduce interpretability \citep{stolfo2025antipodal, smith2025negative}, and when trained on random or out-of-distribution data, SAEs often capture dataset artifacts rather than genuine model-internal patterns \citep{heap2025sparseautoencodersinterpretrandomly, bricken2023monosemanticity}. Such spurious dimensions can be viewed as hallucinated SAE features (henceforth, "Fake Features") that misrepresent the model's true activations.

\begin{figure}[t]
  \centering
  \includegraphics[width=1\linewidth]{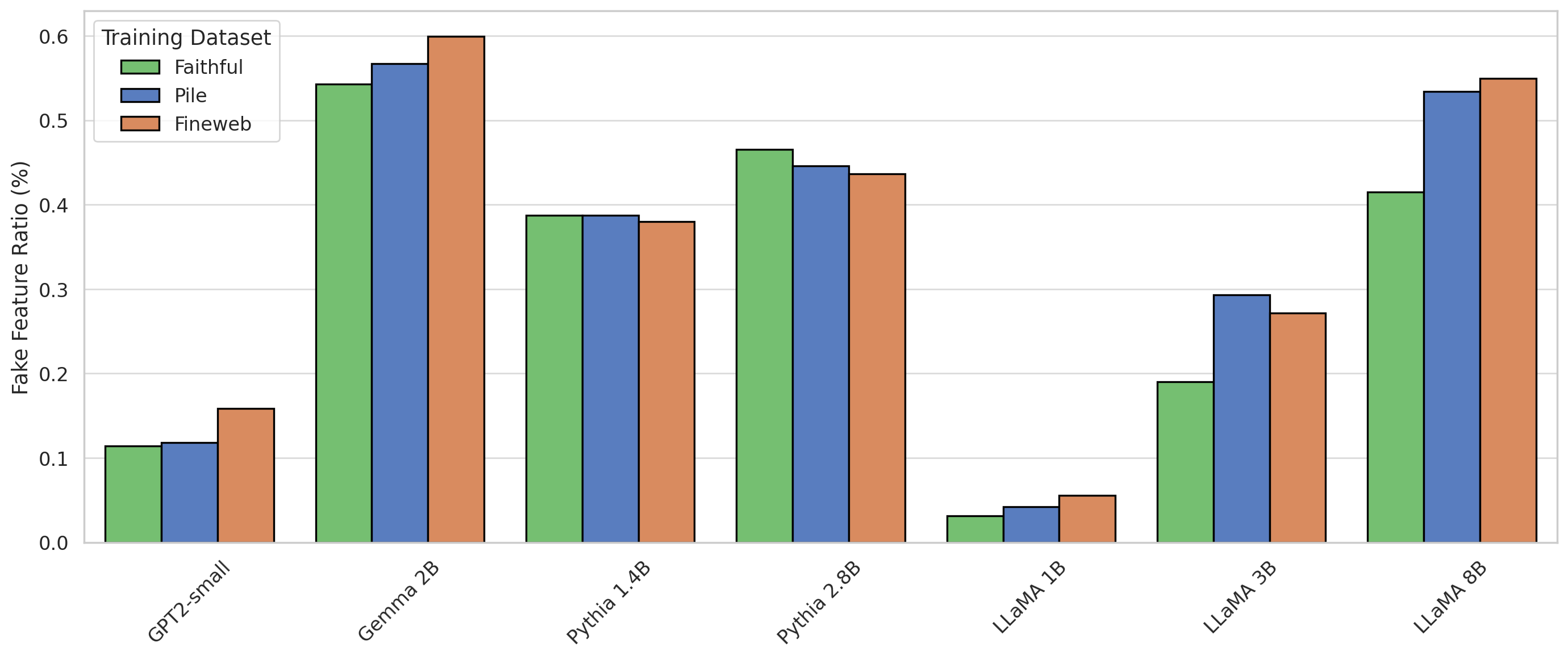}
  \caption{Fake Feature Ratio for SAEs trained on Faithful dataset and Web-based datasets (lower is better). Detailed values can be found in Table \ref{tab:fake-feature}.}
  \label{fig:fake-feature-ratio}
\end{figure}

This work investigates SAE reliability issues, hypothesizing that this unreliability stems from out-of-distribution (OOD) datasets in LLMs \citep{yang-etal-2023-distribution, liu-etal-2024-good}, which are defined as datasets not generalized in LLMs, either absent from pretraining or too complex for the model's capabilities.
To compare the effects of OOD datasets, a Faithful dataset is generated, self-generated synthetic dataset by the LLM, to more accurately reflect LLM-intrinsic features and capabilities.
Faithful SAEs are trained on this dataset and their "faithfulness" is evaluated by measuring reconstruction performance with Cross Entropy (CE), L2 loss, and Explained Variance metrics, while using feature matching techniques \citep{balagansky2025mechanistic, laptev2025analyzefeatureflowenhance, paulo2025sparseautoencoderstraineddata} to assess stability across different seeds.

Based on our experiments, SAEs trained on OOD datasets yield feature sets sensitive to seed differences and lack robustness across different datasets.
First, SAEs were trained on instruction dataset using non-instruction-tuned Pythia \citep{pmlr-v202-biderman23a} models, representing naturally OOD data.
Second, Faithful datasets were compared with potentially OOD Web datasets with different model architectures.
Results showed visible differences in stability across seeds between instruction datasets and Faithful Datasets, while such differences were less pronounced against Web datasets.
Additionally, SAEs trained on Web datasets showed unstable faithfulness across datasets with the above metrics, when compared to FaithfulSAEs.

\section{Background}
\subsection{Mechanistic Interpretability}
Mechanistic Interpretability encompasses approaches that reverse-engineer neural networks through examination of their underlying mechanisms and intermediate representations \citep{olah2020zoom, elhage2021mathematical}.
Researchers systematically analyse multidimensional latent representations, uncovering phenomena such as layer pattern features \citep{olah2017feature, carter2019activation} and neuron-level features \citep{goh2021multimodal, schubert2021high-low} within vision models.
The development of the attention mechanism \citep{vaswani2017attention} and Transformer architecture has intensified research into understanding the emergent capabilities of these models \citep{wei2022emergent}.

\subsection{Superposition Hypothesis}
Within neural networks' representational space, the superposition of word embeddings \citep{arora-etal-2018-linear} has provided substantial evidence for superposition phenomena.
Through studies with toy models, \citealt{elhage2022superposition} elaborated on how the superposition hypothesis emerges via Phase Change in feature dimensionality, establishing connections to compressed sensing \citep{dongho2006compressed, bora2017compressed}.
This hypothesis suggests that polysemanticity emerges as a consequence of neural networks optimizing their representational capacity.
Research has demonstrated that transformer activations contain significant superposition \citep{gurnee2023finding}, suggesting these models encode information as linear combinations of sparse, independent features.

\subsection{Sparse Autoencoders}
Sparse Autoencoders \citep{huben2024sparse, bricken2023monosemanticity} address the Superposition Hypothesis in Transformers by disentangling representational patterns through sparse dictionary learning \citep{olshausen1997sparse, elad2010sparse} for the underlying features.
These models are structured as overcomplete autoencoders, featuring hidden layers with greater dimensionality than their inputs, while incorporating sparsity constraints through $L_1$ regularisation or explicit TopK mechanisms \citep{gao2025scaling}.
Their architectural diversity encompasses various activation functions including ReLU \citep{dunefsky2024transcoders}, JumpReLU \citep{rajamanoharan2025jumping}, TopK \citep{gao2025scaling}, BatchTopK \citep{bussmann2024batchtopk}, alongside different regularisation approaches and decoding mechanisms.

\subsection{SAE Feature}
The SAE features refer to the simplest factorization of hidden activations, which are expected to be human-interpretable latent activations for certain contexts \citep{bricken2023simplest}.
However, sparsity and reconstruction are competing objectives; minimizing loss may occur without preserving conceptual \citep{leask2025sparse} coherence, as sparsity loss randomly suppresses features, which may cause low reproducibility in SAEs.
Moreover, SAEs trained with different seeds or hyperparameters often converge to different sets of features \cite{paulo2025sparseautoencoderstraineddata}.
This instability challenges the assumption that SAEs reliably uncover a unique, model-intrinsic feature dictionary.

\subsection{SAE Weight}

The SAE reconstructs the activations through the following process:

\begin{align}
x_{\text{feature}} &= \sigma(x_{\text{hidden}} \cdot W_{\text{enc}} + b_{\text{enc}}) \\
\hat{x}_{\text{hidden}} &= x_{\text{feature}} \cdot W_{\text{dec}} + b_{\text{dec}}
\end{align}

\text{where } $\sigma$ \text{ is the activation function.}

The encoder weight matrix multiplication can be represented in two forms that yield the same result:

\begin{align}
x_{\text{feature}} &= \sigma\left(\sum_{i=1}^{A} (a_i \cdot w^{\text{enc}}_{i, \cdot}) + b_{\text{enc}}\right) \\
x_{\text{feature}} &= \sigma\left(\bigoplus_{j=1}^{D} (x_{\text{hidden}} \cdot w_{\cdot, j}^{\text{enc}} + b^{\text{enc}}_j)\right)
\end{align}

where $A$ is the activation size and $D$ is the dictionary size and $\bigoplus$ denotes group concatenation.

\begin{itemize}
\item $w_{i,\cdot}^{\text{enc}}$: Each row of the encoder matrix represents the coefficients for linearly disentangling a hidden representation's superposition.
\item $w^{\text{enc}}_{\cdot, j}$: Each column of the encoder matrix represents the coefficients for linearly composing a hidden representation from monosemantic features.
\item $w_{i,j}^{\text{enc}}$: The specific weight at index $(i,j)$ indicates how much the $j$th feature contributes to the superposition at the $i$th hidden representation.
\end{itemize}

The decoder weight matrix multiplication can also be represented in two forms that yield the same result:

\begin{align}
\hat{x}_{\text{hidden}} &= \sum_{j=1}^{D}(d_j\cdot w_{j,\cdot}^{\text{dec}} + b^{\text{dec}}_j) \\
\hat{x}_{\text{hidden}} &= \bigoplus_{i=1}^{A} (x_{\text{feature}} \cdot w^{\text{dec}}_{\cdot, i}) + b_{\text{dec}}
\end{align}

\begin{itemize}
\item $w_{j,\cdot}^{\text{dec}}$: Each row of the decoder matrix shows dictionary features in hidden activations, a Feature Direction \citep{templeton2024scaling} that capture the direction of the feature in the hidden space.
\item $w^{\text{dec}}_{\cdot, i}$: Each column of the decoder matrix shows how each monosemantic dictionary feature contributes to the reconstructed hidden superposition.
\item $w_{j,i}^{\text{dec}}$: The specific weight at index $(j,i)$ specifies how feature $j$ is composited to reconstruct hidden representation $i$.
\end{itemize}
This formulation underscores the critical role of the encoder and decoder weights in disentangling features and accurately reconstructing hidden activations.

\begin{table*}[t]
\centering
\setlength{\tabcolsep}{10pt}
\begin{tabular}{lccccc}
\toprule
\textbf{Model} & \textbf{Total Tokens} & \textbf{Vocab Size} & \textbf{\shortstack{All Token \\ Coverage (\%)}} & \textbf{\shortstack{First Token \\ Coverage (\%)}} & \textbf{\shortstack{KL (Model \\ $\to$ Dataset)}} \\
\midrule
GPT-2 Small    & \num{110718964} & \num{50257}    & 99.80 & 21.49 & 0.2631 \\
Pythia 1.4B    & \num{99999541}  & \num{50254}    & 99.31 & 5.43  & 1.0498 \\
Pythia 2.8B    & \num{103204690} & \num{50254}    & 99.04 & 3.14  & 1.1198 \\
Pythia 6.9B    & \num{57580971}  & \num{50254}    & 99.41 & 13.38 & 0.2893 \\
Gemma 2B       & \num{121006576} & \num{256000}   & 93.44 & 0.40  & 2.2392 \\
LLaMA 3.2-1B   & \num{110070117} & \num{128000}   & 95.78 & 8.27  & 0.1521 \\
LLaMA 3.2-3B   & \num{110395870} & \num{128000}   & 96.09 & 9.18  & 0.1909 \\
LLaMA 3.1-8B   & \num{180268487} & \num{128000}   & 98.04 & 10.31 & 0.1054 \\
\bottomrule
\end{tabular}
\caption{Token statistics across models in the Faithful dataset. KL (Model $\to$ Dataset) represents the forward KL divergence between generated dataset's first token distribution and BOS prediction distribution.}
\label{tab:token-statistics}
\end{table*}

\begin{figure}[t]
  \centering
  \includegraphics[width=\columnwidth]{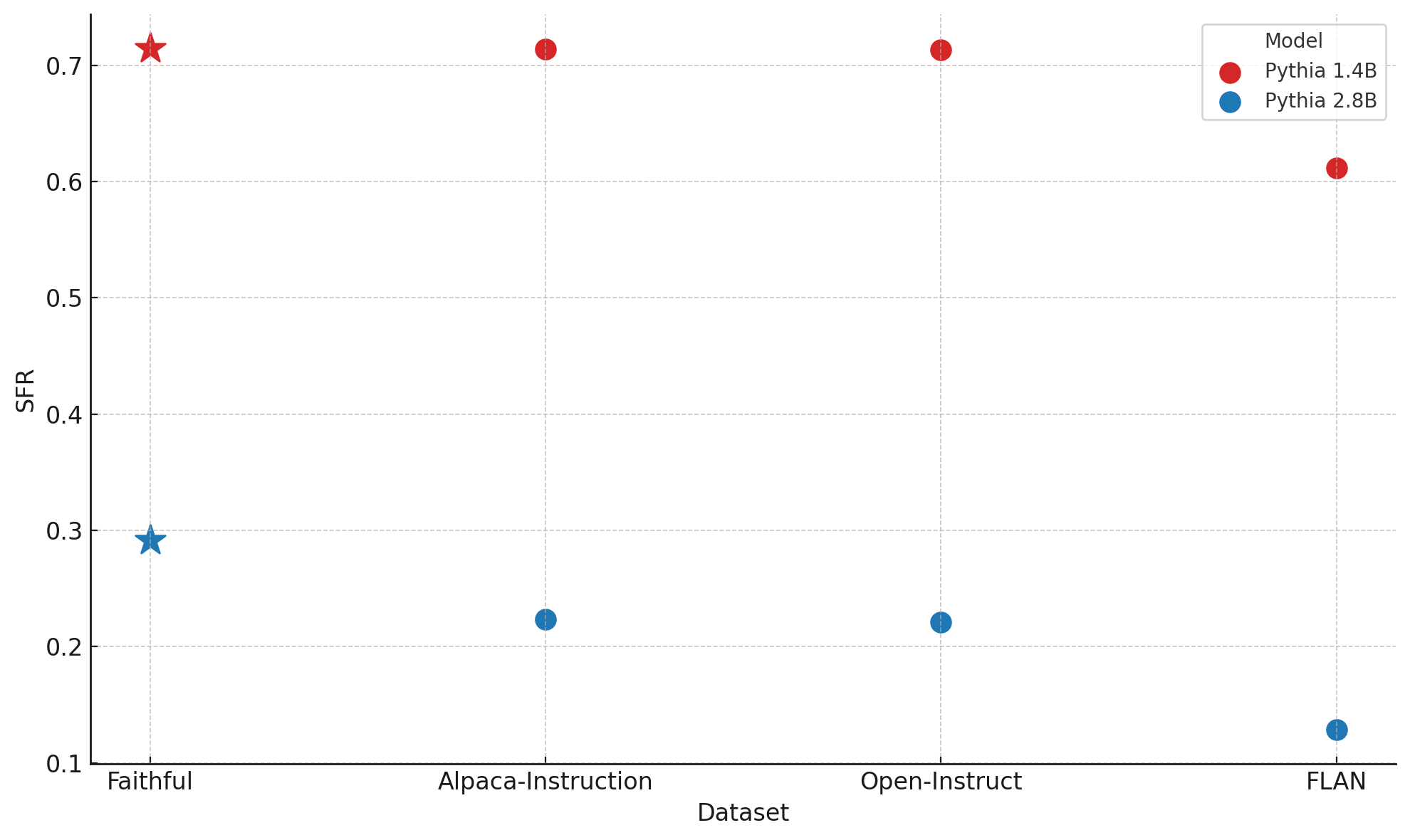}
  \caption{Shared Feature Ratio (SFR) comparison between Faithful Dataset and Instruction Dataset trained SAEs. Detailed values for each run are listed in Table~\ref{tab:shared_feature_ratio_2.8b}.}
  \label{fig:faithful-instruction}
\end{figure}

\section{Methods}

\subsection{Faithful Dataset Generation}

To develop Faithful SAEs that accurately reflect the capabilities of LLMs, the training dataset should closely align with the model's inherent distribution.
The model's generative distribution was captured through unconditional sampling, providing only the Beginning-of-Sequence (BOS) token as the input prompt.
This is referred to as the Faithful Dataset, as it directly corresponds to the model's natural next-token prediction distribution.

\subsection{Faithful SAE Training}
Using the generated Faithful Dataset, the Top-K SAEs \citep{gao2025scaling} were trained.
To demonstrate the faithfulness of the trained models, two Faithful SAEs were trained with the same configuration but different seeds.
For comparison, SAEs with the same seeds were also trained using not only the SAE dataset but also various other datasets.

\subsection{Evaluation Metrics}
Faithfulness was evaluated by examining individual learned features in the SAE latent space across different seeds, with specific metrics as follows.
To quantify the faithfulness of SAEs, several complementary metrics were employed.
The primary metrics include Shared Feature Ratio, Cross-Entropy (CE) difference, L2 reconstruction error, and Explained Variance. 

\subsection{Feature Matching}
To understand how different training conditions affect the learned representations within SAEs, features discovered by different SAEs are compared using Feature Matching \citep{balagansky2025mechanistic, laptev2025analyzefeatureflowenhance, paulo2025sparseautoencoderstraineddata}.
A common approach, inspired by Maximum Marginal Cosine Similarity (MMCS) \citep{sharkey2022sparse}, computes the cosine similarity between feature vectors using their corresponding decoder weight vectors, where $w_j = w^{dec}_{j, \cdot}$.

\[m_{j} = \max_{w_{k}' \in W_2} \frac{w_{j} \cdot w_{k}'}{\|w_{j}\| \, \|w_{k}'\|}\]

Following \citet{paulo2025sparseautoencoderstraineddata}, the Hungarian matching algorithm \citep{Kuhn1955TheHM} was used to find an optimal one-to-one correspondence between feature sets. We compute the similarity matrix $S \in \mathbf{R}^{d \times d}$ between all features of two SAEs:

\[S_{j,k} = \frac{w^{dec}_{j, \cdot} \cdot w^{dec'}_{k, \cdot}}{\|w^{dec}_{j, \cdot}\| \, \|w^{dec'}_{k, \cdot}\|}\]

After applying the Hungarian algorithm to find the optimal assignment that maximizes the total similarity, each feature is classified based on a threshold $\tau_s$ into 'shared' or 'orphan' features, terminology introduced by \citet{paulo2025sparseautoencoderstraineddata}:

\[\text{Feature Type}(d_j) = \begin{cases} \text{shared} & \text{if } S_{j,k} \ge \tau_s, \\ \text{orphan} & \text{if } S_{j,k} < \tau_s. \end{cases}\]

This approach ensures that each feature from one SAE is matched with at most one feature from the other SAE, providing a measure of feature set similarity.

Using this methodology, the Shared Feature Ratio is defined as the proportion of shared features relative to the total number of features in an SAE:

\[SFR = \frac{|\{d_j \in D \mid S_{j,k} \geq \tau_s\}|}{|D|}\]

where $D$ is the complete dictionary of features in the SAE, and $|\cdot|$ denotes the cardinality of a set.

\subsection{Fake Feature Ratio}
Frequently activating features have been identified as problematic in SAE literature \citep{stolfo2025antipodal, smith2025negative}, often leading to poor interpretability.
"Fake Feature" is defined as a feature that activate on randomly generated token sequences (OOD inputs). 
A feature is considered fake if it frequently activates on more than a certain threshold $\tau_f$ of OOD samples. The Fake Feature Ratio (FFR) is defined as:

\[\text{FFR} = \frac{|\{i \in D : \text{activation frequency}(i) > \tau_f\}|}{|D|}\]
where $D$ is the total feature dictionary. Lower FFR indicates better feature quality.

\subsection{SAE Probing}
To evaluate downstream task performance of SAE, three approaches are compared on classification tasks: original model activations (Baseline), sparse feature activations (SAE), and reconstructed activations (Reconstruction). Logistic regression probes are trained for each representation type and accuracy and F1 scores are measured across SST-2, CoLA, AG News, and Yelp Polarity datasets. A faithful SAE should show minimal performance drop between baseline and SAE/reconstruction approaches.

\section{Experiments}

We used SFR with threshold $\tau_s$ as 0.7 between SAEs trained with different random seeds.
For the FFR threshold, we followed \citet{smith2025negative} and set $\tau_f = 0.1$.
For each experiment, we trained multiple SAEs using two different initialization seeds while keeping all other hyperparameters constant.
For all datasets except LLaMA 8B, we used 100M tokens for training. For LLaMA 8B, we used 150M tokens to ensure convergence. FFR measurement was measured by generating 1M tokens and averaged across all different seed SAEs for a reliable measure. 

\subsection{Instruction Dataset Comparison}

The training dataset used during pre-training must be publicly available. For example, models like LLaMA \cite{grattafiori2024llama3herdmodels} do not disclose their training data.
The research leveraged the fact that pre-trained models have internalised the distribution of their training data and rely on this distribution for inference. Therefore, the pre-trained model was treated as a proxy for its training distribution and used to generate synthetic data.
The open-source Pythia \cite{pmlr-v202-biderman23a} model was employed, for which the training dataset is publicly available.

For the Out-of-Distribution (OOD) datasets, Instruction Tuning \citep{wei2022finetuned} datasets were used: FLAN \citep{pmlr-v202-longpre23a}, OpenInstruct \citep{wang2023how}, and Alpaca dataset \citep{taori_alpaca_2023}. Selecting an uncensored dataset was crucial for constructing a valid OOD benchmark. This decision was based on the fact that commonly used datasets for training SAEs contain data scraped from the same sources. Additionally, models with different parameter scales were compared: Pythia 1.4B and Pythia 2.8B, to study the impact of model size on SAE faithfulness.

\begin{figure}[t]
  \centering
  \includegraphics[width=\columnwidth]{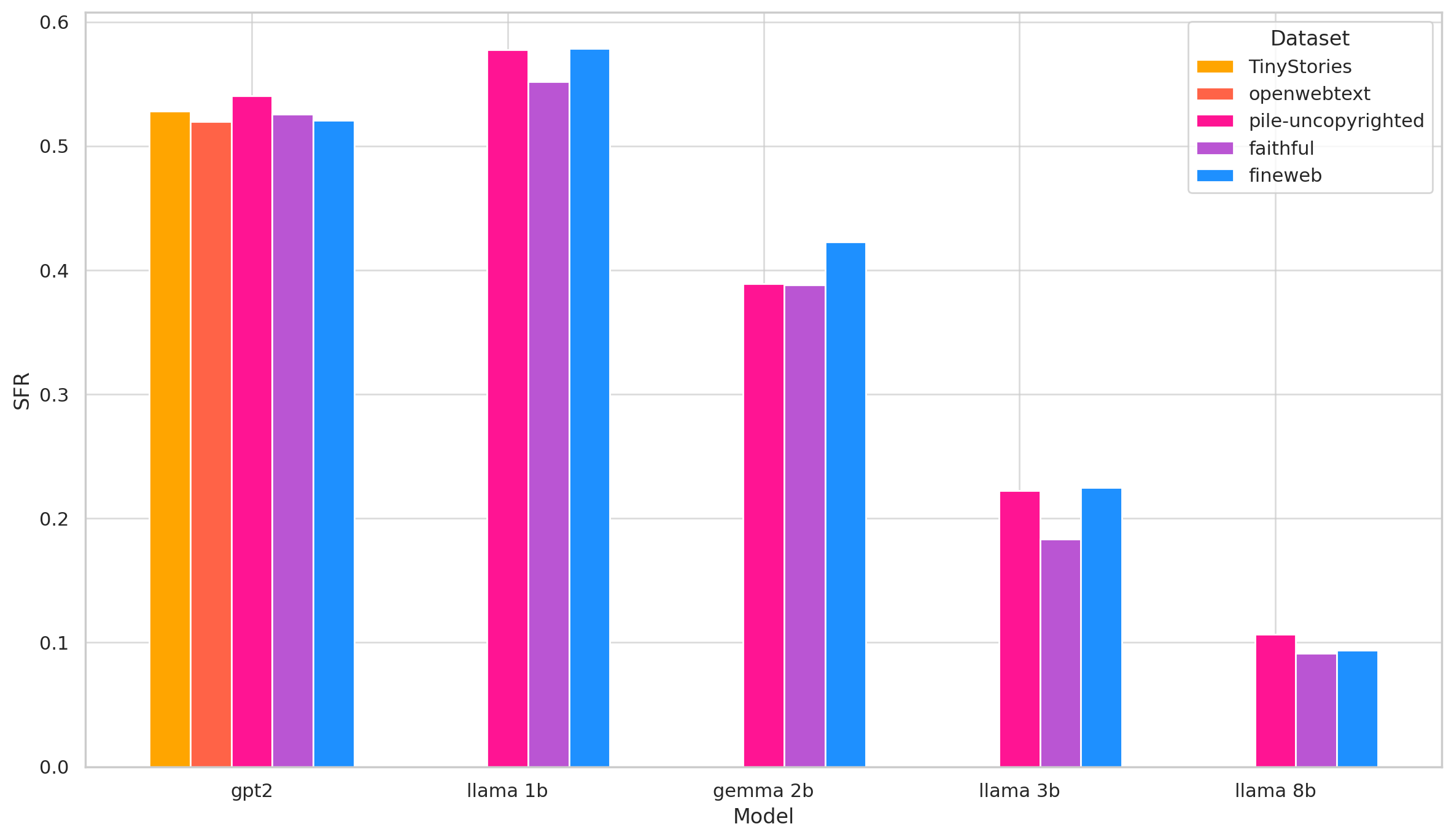}
  \caption{Shared Feature Ratio by model and dataset. 
SAE training hyperparameters are listed in Appendix~\ref{appendix:sae-training-hyperparameters}, 
and complete results appear in Table~\ref{tab:feature-matching}.}
  \label{fig:feature-matching}
\end{figure}

\subsection{Web-based Dataset Comparison}
\label{subsec:web-based-dataset}

For cross-architecture comparison against Web-based dataset and Faithful dataset, the Top-K SAE model \citep{gao2025scaling} was utilized. 
To evaluate a diverse range of architectures and examine scaling effects, five models were employed: GPT-2 Small \citep{radford2019language}, LLaMA 3.2 1B, LLaMA 3.2 3B, LLaMA 3.1 8B \citep{grattafiori2024llama3herdmodels}, and Gemma 2B \citep{gemmateam2024gemma2improvingopen}.
SAEs were trained on three distinct datasets—The Pile \citep{gao2021pile}, FineWeb \citep{penedo2024the}, and our Faithful Dataset—for each model architecture, with hyperparameters specified in Table \ref{tab:file-configurations}.
After training SAEs across different datasets and architectures using two initialization seeds, the SFR metric was compared when only the seed was altered to assess model stability.

\subsection{SAE Faithfulness Metrics}

The objective is to determine whether training SAEs on the generated Faithful dataset produces more faithful sparse representations of model activations. 
It is argued that a more faithful SAE should adapt more flexibly to the model when encoding and decoding activations, maintaining the essential information flow through the model.
To quantify this faithfulness, Cross-Entropy (CE) difference, L2 reconstruction error, and Explained Variance were used as proxy metrics, comparing trained SAEs to measure their impact on the underlying model.
This evaluation was conducted using SAEs trained on The Pile, FineWeb, and the Faithful Dataset, and extended the test suite to include not only these three datasets but also OpenWebText \citep{gokaslan2019openwebtext} and TinyStories \citep{li2024tinystories} for comprehensive assessment.

\subsection{SAE Probing}

For our SAE Probing experiments, four diverse classification datasets were selected: SST-2 \citep{socher-etal-2013-recursive}, CoLA \citep{warstadt-etal-2019-neural}, AG News and Yelp Polarity \citep{NIPS2015_250cf8b5}.
For each dataset, reconstructed activations were used as input for logistic regression classifier.
Activations were aggregated by mean pooling on every token in the sequence.
The classifiers were trained on each representation type and accuracy score was measured, using a maximum of 100,000 samples for training.
The accuracy scores were averaged across all seed SAEs to obtain more reliable data.

\begin{figure*}[t]
  \centering
  \includegraphics[width=0.7\textwidth]{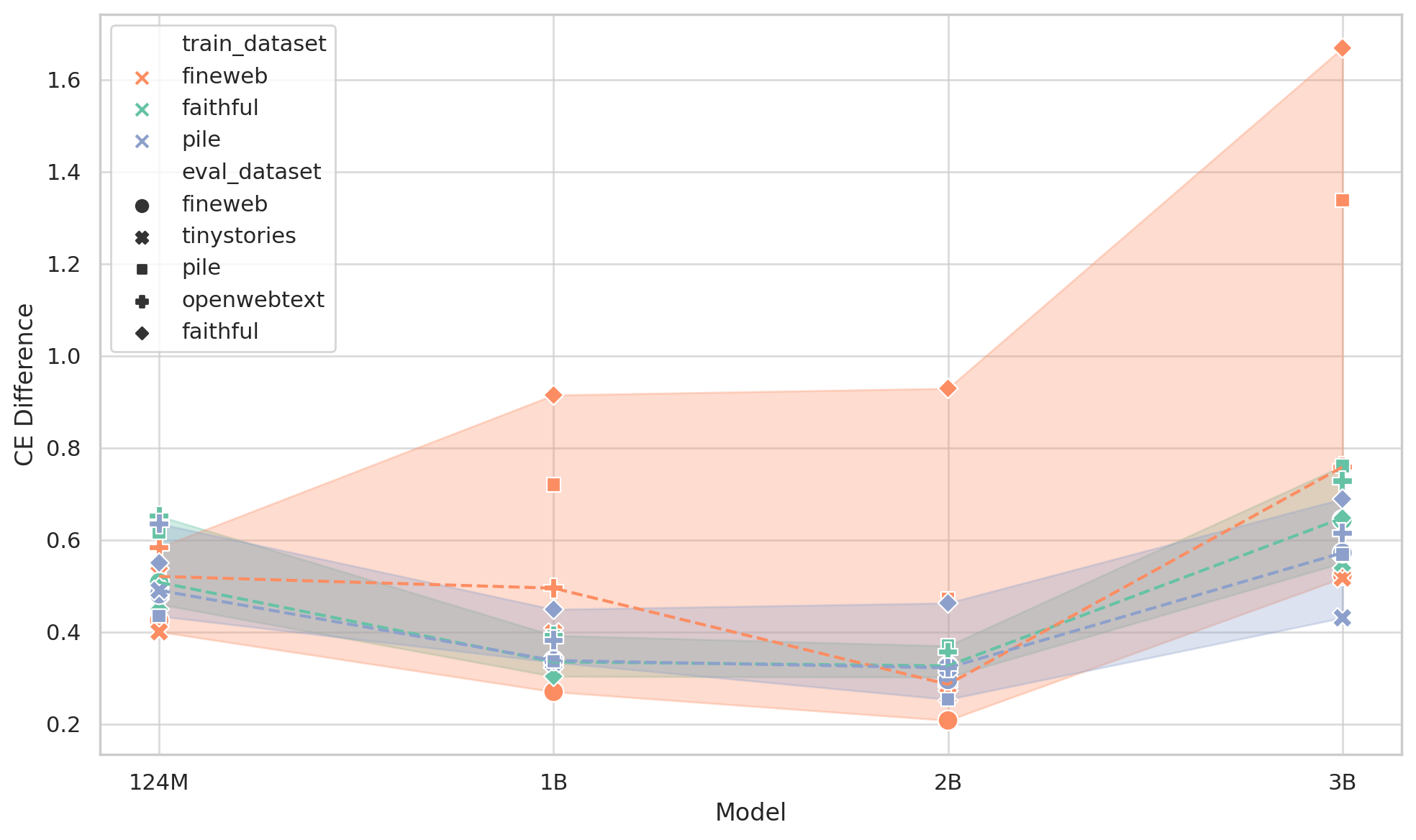}
  \caption{Cross-Entropy difference between SAEs trained on different datasets. Colors represent training datasets: orange for FineWeb, gray for Pile-Uncopyrighted, and green for Faithful dataset.
  Point shapes indicate evaluation datasets: circles for FineWeb, squares for The Pile, X markers for TinyStories, crosses for OpenWebText, and diamonds for Faithful dataset.
  You can find the detailed metrics in Appendix \ref{sec:faithful-saes}.}
  \label{fig:ce-diff}
\end{figure*}

\section{Results}

\subsection{Impact of OOD Levels on SAE Stability Across Datasets}

As shown in Table \ref{tab:shared_feature_ratio_2.8b}, FaithfulSAEs, trained on a synthetic dataset, exhibit greater stability across seeds compared to SAEs trained on mixed or instruction-based datasets. These results support our hypothesis that higher OOD levels reduce SFR. Notably, layer 16 demonstrates higher stability than layer 8, likely due to SAEs capturing more complex features in deeper layers.

\begin{table}[!ht]
    \centering
    \begin{tabular}{lcc}
    \toprule
    Dataset & Pythia 1.4B & Pythia 2.8B \\
    \midrule
    Faithful & \textbf{0.7145} & \textbf{0.2911} \\
    Alpaca-Instruction & 0.7138 & 0.2231 \\
    Open-Instruct & 0.7134 & 0.2210 \\
    FLAN & 0.6113 & 0.1283 \\
    \bottomrule
    \end{tabular}
    \caption{Shared Feature Ratio for Pythia 1.4B and 2.8B model. AI denotes Alpaca-Instruction for compactness.}
    \label{tab:shared_feature_ratio_2.8b}
\end{table}

\subsection{SFR on Cross-Model Synthetic Datasets}

\begin{table}[!ht]
\centering
\begin{tabular}{llll}
\toprule
Target Model & Source Model & SFR \\
\midrule
Pythia 2.8b & Pythia 2.8b & \textbf{0.2911} \\
Pythia 2.8B & Pythia 1.4B & 0.2288 \\

\midrule

Pythia 1.4B & Pythia 1.4B & \textbf{0.7145} \\
Pythia 1.4B & Pythia 2.8B & 0.6887 \\
\bottomrule
\end{tabular}
\caption{Shared Feature Ratio on Pythia models. FaithfulSAEs were trained on target models with synthetic datasets generated from source models.}
\label{tab:cross_model_seed_similarity}
\end{table}

From Table \ref{tab:cross_model_seed_similarity}, we observe that SFR is consistently higher when the target model is the same as the source model (e.g., training SAEs on a Pythia 2.8B model with a synthetic dataset from a 2.8B model), and lower when the source and target models are different.
This suggests that SAE training on its own synthetic dataset is more stable even within the same model family trained on the same dataset with different scaling.
This indicates that SFR differences stem from out-of-distribution effects, and a smaller model's dataset is not necessarily easier to learn stable feature sets from.
The results are consistent with our hypothesis: more OOD input leads to lower SAE stability across seeds (lower SFR), while less OOD leads to more consistent SAE training (higher SFR).

\begin{figure*}[t]
  \centering
  \includegraphics[width=1\linewidth]{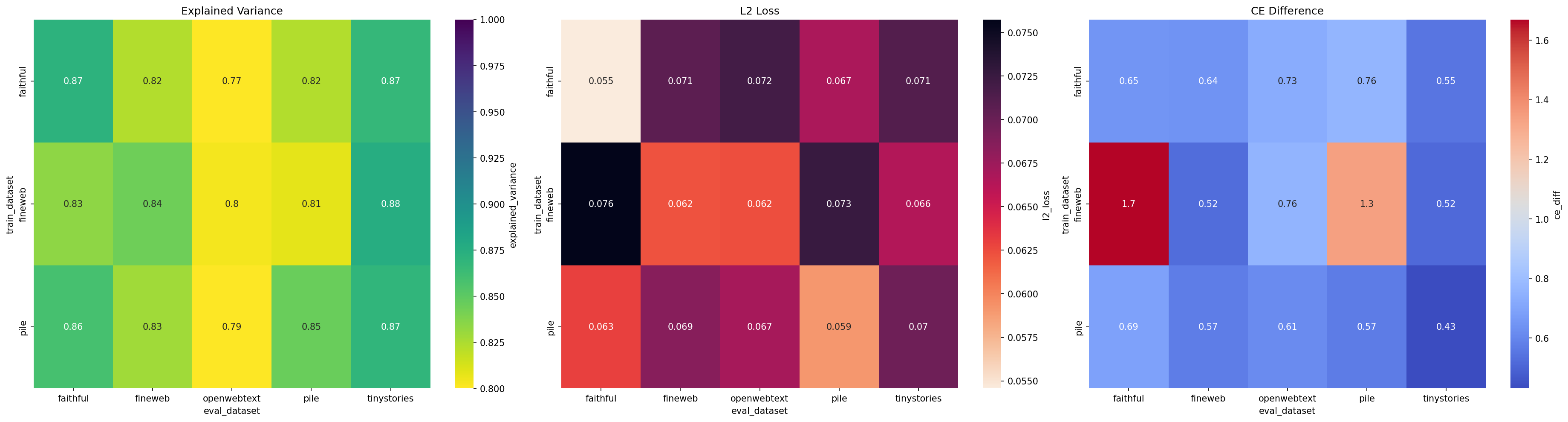}
  \caption{Faithful SAE representation for LLaMa 8B.
  This figure shows the SAE's reconstruction of the LLaMa 8B hidden state and its faithfulness across datasets.}
  \label{fig:sae-llama8}
\end{figure*}

\subsection{Performance on Web-based Datasets}

The Faithful dataset did not demonstrate higher SFR compared to web-based datasets as shown in Figure \ref{fig:feature-matching}; rather, it showed lower SFR across most models. As evident in Table \ref{tab:feature-matching}, the Faithful dataset exhibited lower SFR than FineWeb or The Pile for all models. 

\begin{table}[!ht]
  \small
  \centering
  \setlength{\tabcolsep}{10pt}
  \begin{tabular}{lccc}
  \toprule
  \textbf{Model} & \textbf{Pile} & \textbf{Faithful} & \textbf{FineWeb} \\
  \midrule
  GPT-2         & \textbf{0.5405} & 0.5258 & 0.5209 \\
  LLaMA 1B  & 0.5778 & 0.5517 & \textbf{0.5789} \\
  Gemma 2B      & 0.3889 & 0.3881 & \textbf{0.4229} \\
  LLaMA 3B  & 0.2222 & 0.1835 & \textbf{0.2248} \\
  LLaMA 8B  & \textbf{0.1066} & 0.0914 & 0.0936 \\
  \bottomrule
  \end{tabular}
  \caption{Shared Feature Ratio across models and datasets. It compares SAEs trained with identical settings but different seeds. The models listed were used for SAE activation extraction, and the datasets on the right were used for training them.}
  \label{tab:feature-matching}
\end{table}

We concluded that this issue arises because web-based datasets are sufficiently diverse to encompass model coverage, and out-of-distribution data beyond the scope of the Faithful dataset does not negatively impact the robustness of SAEs.

By observing that GPT2 relatively showed similar SFR with other Web-based datasets, while the larger models such as Gemma and LLaMA consistently showed lower SFR.
This is because the pretraining datasets of Gemma and LLaMA already contain Web-based data generalization, which means they are not OOD datasets.
To address this limitation, generating larger Faithful datasets would better cover the full range of model capabilities, which we analyze in more detail in Subsection \ref{subsec:results-faithfulness} by comparing SAE faithfulness.

\subsection{Faithfulness of Faithful Dataset}
\label{subsec:results-faithfulness}

As shown in Table \ref{tab:token-statistics}, KL divergence values stay below 2 except for Gemma 2B, demonstrating effective mode covering via Forward KL.
The table confirms >90\% Unique Tokens Used in All Positions, indicating adequate model distribution capture.
However, first token distribution lacks vocabulary breadth, possibly explaining why Figure \ref{fig:feature-matching} shows FaithfulSAEs underperforming Web-based SAEs.
Alternative approaches include starting with a flat distribution instead of BOS tokens or increasing the sampling temperature.

In Appendix \ref{subsec:faithful-dataset}, we verify the proper generation of the dataset by confirming that the distribution of top tokens follows the predicted distribution of BOS tokens. However, due to limited sampling in the dataset, it does not cover all token distributions from the BOS prediction, which follow a logarithmic decrease.

\subsection{Faithfulness of FaithfulSAE}
\label{subsec:faithfulness-sae}

To determine whether training SAEs on the generated Faithful dataset produces more faithful SAEs, we evaluated model fidelity during activation encoding and decoding processes with trained SAEs as presented in Table \ref{tab:file-configurations}.
We measured Cross-Entropy difference, L2, and Explained Variance metrics across five datasets.
The full results are available in Appendix \ref{sec:faithful-saes}, while the results for LLaMa 8B are shown in Figure \ref{fig:sae-llama8}.

Although FineWeb SAE showed higher SFR than Faithful SAE, it demonstrated significantly higher CE difference and overall lower generalized performance on faithfulness metrics.
SAEs trained on The Pile achieved higher SFR, while faithfulness metrics were similar as shown in Appendix \ref{sec:faithful-saes}.
SAEs trained exclusively on the Faithful Dataset demonstrated more stable performance across multiple evaluation datasets compared to FineWeb.

\subsection{SAE Probing}

\begin{figure*}[t]
  \centering
  \includegraphics[width=1\linewidth]{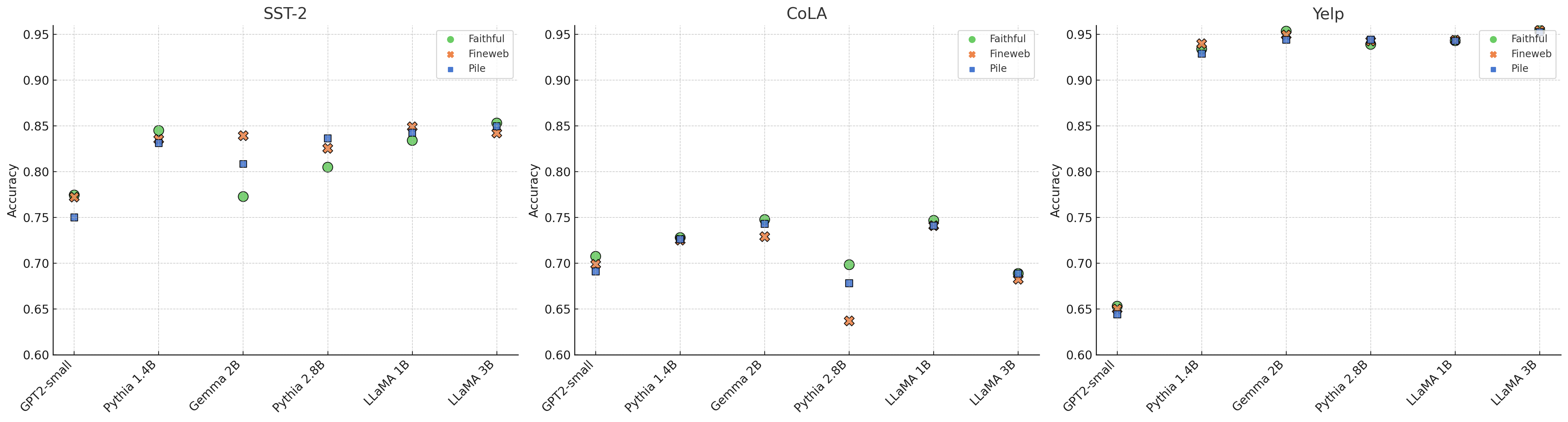}
  \caption{SAE Probing performance comparison between FaithfulSAE and Web-based SAEs with different types of LLM architectures. Detailed values can be found in Table \ref{tab:sae-probing-table}.}
  \label{fig:sae-probing}
\end{figure*}

Notably in Figure \ref{fig:sae-probing}, FaithfulSAE demonstrates overall better performance compared to the other Web-based trained SAEs.
FaithfulSAE achieved superior performance in 12 out of 18 cases across six models and three classification tasks. While performance varied by task, FaithfulSAE consistently outperformed alternatives on the CoLA dataset across all model configurations.
Despite showing lower SFR compared to Web-based datasets, the higher downstream task performance of FaithfulSAE suggests it more accurately reflects the model's hidden state with less reconstruction noise.

\subsection{Fake Feature}

While FaithfulSAE generally shows lower SFR compared to web-based datasets, it demonstrates better performance in terms of FFR (lower), suggesting potential benefits for interpretability with the Faithful Dataset.
Among the 7 models tested, 5 models showed lower FFR with FaithfulSAE, with the exception of the Pythia model family.
This is likely because the Pythia model, as mentioned above, was trained exclusively on The Pile dataset, which closely overlaps with the web-based FineWeb and The Pile datasets used for comparison.
We also observed that within the same model family, larger models showed higher FFR with FaithfulSAE, indicating that interpretability becomes more challenging as model size increases.

\section{Conclusion}

Out-of-distribution datasets that exceed a model's pretraining distribution or capabilities hinder SAEs from reliably identifying consistent feature sets across different initialization seeds. 
To mitigate this, we proposed Faithful SAE—trained on the model's own synthetic dataset—to ensure that training remains strictly within the model's inherent capabilities.
Our experiments showed that FaithfulSAEs yield higher SFR than those trained on instruction-tuned datasets and outperform SAEs trained on Web-based datasets in the SAE proving task.
While FaithfulSAEs obtain lower FFR than web-based dataset trained SAEs leading to improved potential interpretability, they also offer a key advantage: encapsulation.



\section{Limitations}

While Faithful Datasets improve feature consistency for non-instruction-tuned models, our experiment lacked evaluation on instruction-tuned or reasoning models.
Our evaluation of Shared Feature Ratio may not fully reflect the complexity of high-dimensional feature spaces, and we did not assess the interpretability of individual features. 
Specifically, Shared Feature Ratio was higher compared to instruction datasets, but lower compared to web-based datasets.
Additionally, we need to verify whether Faithful SAE provides interpretable explanations for individual features through case studies.
Although we defined the Fake Feature Ratio and confirmed lower values, we did not remove these features or assess their interpretability further.

\section{Future Work}

This work shows that our approach can reduce Fake Features and improve probing performance.
An important direction for future research is exploring improved dataset generation and training strategies that could completely outperform Web-based methods.
Such progress would further validate the promise of training interpretability models using only the model itself, without reliance on external data. 
This dataset independence could be particularly advantageous for interpretability in domain-specific generative models where data is scarce.
For example, the FaithfulSAE approach could be adopted for interpretability of models in biology or robotics where data production costs are high.

Another priority is to evaluate whether Faithful SAEs provide meaningful and interpretable explanations for individual features through detailed case studies.
For example, we hypothesize that pruning Fake Features from a Faithful SAE may yield a representation close to the Simplest Factorization \citep{bricken2023simplest}, aligning with the principle of Minimal Description Length \citep{ayonrinde2024interpretability}.
Confirming this connection remains an open and exciting avenue for future investigation.



\bibliography{custom}

\begin{thebibliography}{51}
\providecommand{\natexlab}[1]{#1}

\bibitem[{Arora et~al.(2018)Arora, Li, Liang, Ma, and Risteski}]{arora-etal-2018-linear}
Sanjeev Arora, Yuanzhi Li, Yingyu Liang, Tengyu Ma, and Andrej Risteski. 2018.
\newblock \href {https://doi.org/10.1162/tacl_a_00034} {Linear algebraic structure of word senses, with applications to polysemy}.
\newblock \emph{Transactions of the Association for Computational Linguistics}, 6:483--495.

\bibitem[{Ayonrinde et~al.(2024)Ayonrinde, Pearce, and Sharkey}]{ayonrinde2024interpretability}
Kola Ayonrinde, Michael~T. Pearce, and Lee Sharkey. 2024.
\newblock \href {https://arxiv.org/abs/2410.11179} {Interpretability as compression: Reconsidering sae explanations of neural activations with mdl-saes}.
\newblock \emph{Preprint}, arXiv:2410.11179.

\bibitem[{Balagansky et~al.(2025)Balagansky, Maksimov, and Gavrilov}]{balagansky2025mechanistic}
Nikita Balagansky, Ian Maksimov, and Daniil Gavrilov. 2025.
\newblock \href {https://openreview.net/forum?id=MDvecs7EvO} {Mechanistic permutability: Match features across layers}.
\newblock In \emph{The Thirteenth International Conference on Learning Representations}.

\bibitem[{Biderman et~al.(2023)Biderman, Schoelkopf, Anthony, Bradley, O'Brien, Hallahan, Khan, Purohit, Prashanth, Raff, Skowron, Sutawika, and Van Der~Wal}]{pmlr-v202-biderman23a}
Stella Biderman, Hailey Schoelkopf, Quentin~Gregory Anthony, Herbie Bradley, Kyle O'Brien, Eric Hallahan, Mohammad~Aflah Khan, Shivanshu Purohit, Usvsn~Sai Prashanth, Edward Raff, Aviya Skowron, Lintang Sutawika, and Oskar Van Der~Wal. 2023.
\newblock \href {https://proceedings.mlr.press/v202/biderman23a.html} {Pythia: A suite for analyzing large language models across training and scaling}.
\newblock In \emph{Proceedings of the 40th International Conference on Machine Learning}, volume 202 of \emph{Proceedings of Machine Learning Research}, pages 2397--2430. PMLR.

\bibitem[{Bora et~al.(2017)Bora, Jalal, Price, and Dimakis}]{bora2017compressed}
Ashish Bora, Ajil Jalal, Eric Price, and Alexandros~G Dimakis. 2017.
\newblock \href {https://proceedings.mlr.press/v70/bora17a/bora17a.pdf} {Compressed sensing using generative models}.
\newblock In \emph{Proceedings of the 34th International Conference on Machine Learning (ICML)}, volume~70 of \emph{Proceedings of Machine Learning Research}, pages 537--546, Sydney, Australia. PMLR.

\bibitem[{Bricken et~al.(2023{\natexlab{a}})Bricken, Batson, Templeton, Jermyn, Henighan, and Olah}]{bricken2023simplest}
Trenton Bricken, Joshua Batson, Adly Templeton, Adam Jermyn, Tom Henighan, and Chris Olah. 2023{\natexlab{a}}.
\newblock \href {https://transformer-circuits.pub/2023/may-update/index.html#simple-factorization} {Features as the simplest factorization}.
\newblock Part of the May 2023 Circuits Updates by the Anthropic interpretability team.

\bibitem[{Bricken et~al.(2023{\natexlab{b}})Bricken, Templeton, Batson, Chen, Jermyn, Conerly, Turner, Anil, Denison, Askell, Lasenby, Wu, Kravec, Schiefer, Maxwell, Joseph, Hatfield-Dodds, Tamkin, Nguyen, McLean, Burke, Hume, Carter, Henighan, and Olah}]{bricken2023monosemanticity}
Trenton Bricken, Adly Templeton, Joshua Batson, Brian Chen, Adam Jermyn, Tom Conerly, Nick Turner, Cem Anil, Carson Denison, Amanda Askell, Robert Lasenby, Yifan Wu, Shauna Kravec, Nicholas Schiefer, Tim Maxwell, Nicholas Joseph, Zac Hatfield-Dodds, Alex Tamkin, Karina Nguyen, and 6 others. 2023{\natexlab{b}}.
\newblock \href {https://transformer-circuits.pub/2023/monosemantic-features/index.html} {Towards monosemanticity: Decomposing language models with dictionary learning}.
\newblock \emph{Transformer Circuits Thread}.

\bibitem[{Bussmann et~al.(2024)Bussmann, Leask, and Nanda}]{bussmann2024batchtopk}
Bart Bussmann, Patrick Leask, and Neel Nanda. 2024.
\newblock \href {https://openreview.net/forum?id=d4dpOCqybL} {Batchtopk sparse autoencoders}.
\newblock In \emph{NeurIPS 2024 Workshop on Scientific Methods for Understanding Deep Learning}.

\bibitem[{Carter et~al.(2019)Carter, Armstrong, Schubert, Johnson, and Olah}]{carter2019activation}
Shan Carter, Zan Armstrong, Ludwig Schubert, Ian Johnson, and Chris Olah. 2019.
\newblock \href {https://doi.org/10.23915/distill.00015} {Activation atlas}.
\newblock \emph{Distill}.

\bibitem[{Donoho(2006)}]{dongho2006compressed}
D.L. Donoho. 2006.
\newblock \href {https://doi.org/10.1109/TIT.2006.871582} {Compressed sensing}.
\newblock \emph{IEEE Transactions on Information Theory}, 52(4):1289--1306.

\bibitem[{Dunefsky et~al.(2024)Dunefsky, Chlenski, and Nanda}]{dunefsky2024transcoders}
Jacob Dunefsky, Philippe Chlenski, and Neel Nanda. 2024.
\newblock \href {https://openreview.net/forum?id=J6zHcScAo0} {Transcoders find interpretable {LLM} feature circuits}.
\newblock In \emph{The Thirty-eighth Annual Conference on Neural Information Processing Systems}.

\bibitem[{Elad(2010)}]{elad2010sparse}
Michael Elad. 2010.
\newblock \href {https://doi.org/10.1007/978-1-4419-7011-4} {\emph{Sparse and Redundant Representations: From Theory to Applications in Signal and Image Processing}}, 1 edition.
\newblock Springer New York.

\bibitem[{Elhage et~al.(2022)Elhage, Hume, Olsson, Schiefer, Henighan, Kravec, Hatfield-Dodds, Lasenby, Drain, Chen, Grosse, McCandlish, Kaplan, Amodei, Wattenberg, and Olah}]{elhage2022superposition}
Nelson Elhage, Tristan Hume, Catherine Olsson, Nicholas Schiefer, Tom Henighan, Shauna Kravec, Zac Hatfield-Dodds, Robert Lasenby, Dawn Drain, Carol Chen, Roger Grosse, Sam McCandlish, Jared Kaplan, Dario Amodei, Martin Wattenberg, and Christopher Olah. 2022.
\newblock \href {https://transformer-circuits.pub/2022/toy\_model/index.html} {Toy models of superposition}.
\newblock \emph{Transformer Circuits Thread}.

\bibitem[{Elhage et~al.(2021)Elhage, Nanda, Olsson, Henighan, Joseph, Mann, Askell, Bai, Chen, Conerly, DasSarma, Drain, Ganguli, Hatfield-Dodds, Hernandez, Jones, Kernion, Lovitt, Ndousse, Amodei, Brown, Clark, Kaplan, McCandlish, and Olah}]{elhage2021mathematical}
Nelson Elhage, Neel Nanda, Catherine Olsson, Tom Henighan, Nicholas Joseph, Ben Mann, Amanda Askell, Yuntao Bai, Anna Chen, Tom Conerly, Nova DasSarma, Dawn Drain, Deep Ganguli, Zac Hatfield-Dodds, Danny Hernandez, Andy Jones, Jackson Kernion, Liane Lovitt, Kamal Ndousse, and 6 others. 2021.
\newblock \href {https://transformer-circuits.pub/2021/framework/index.html} {A mathematical framework for transformer circuits}.
\newblock \emph{Transformer Circuits Thread}.

\bibitem[{Faruqui et~al.(2015)Faruqui, Tsvetkov, Yogatama, Dyer, and Smith}]{faruqui-etal-2015-sparse}
Manaal Faruqui, Yulia Tsvetkov, Dani Yogatama, Chris Dyer, and Noah~A. Smith. 2015.
\newblock \href {https://doi.org/10.3115/v1/P15-1144} {Sparse overcomplete word vector representations}.
\newblock In \emph{Proceedings of the 53rd Annual Meeting of the Association for Computational Linguistics and the 7th International Joint Conference on Natural Language Processing (Volume 1: Long Papers)}, pages 1491--1500, Beijing, China. Association for Computational Linguistics.

\bibitem[{Gao et~al.(2021)Gao, Biderman, Black, Golding, Hoppe, Foster, Phang, He, Thite, Nabeshima, Presser, and Leahy}]{gao2021pile}
Leo Gao, Stella Biderman, Sid Black, Laurence Golding, Travis Hoppe, Charles Foster, Jason Phang, Horace He, Anish Thite, Noa Nabeshima, Shawn Presser, and Connor Leahy. 2021.
\newblock \href {https://arxiv.org/abs/2101.00027} {The pile: An 800gb dataset of diverse text for language modeling}.
\newblock \emph{CoRR}, abs/2101.00027.

\bibitem[{Gao et~al.(2024)Gao, la~Tour, Tillman, Goh, Troll, Radford, Sutskever, Leike, and Wu}]{gao2025scaling}
Leo Gao, Tom~Dupre la~Tour, Henk Tillman, Gabriel Goh, Rajan Troll, Alec Radford, Ilya Sutskever, Jan Leike, and Jeffrey Wu. 2024.
\newblock \href {https://openreview.net/forum?id=tcsZt9ZNKD} {Scaling and evaluating sparse autoencoders}.
\newblock In \emph{The Thirteenth International Conference on Learning Representations}.

\bibitem[{Goh et~al.(2021)Goh, †, †, Carter, Petrov, Schubert, Radford, and Olah}]{goh2021multimodal}
Gabriel Goh, Nick~Cammarata †, Chelsea~Voss †, Shan Carter, Michael Petrov, Ludwig Schubert, Alec Radford, and Chris Olah. 2021.
\newblock \href {https://doi.org/10.23915/distill.00030} {Multimodal neurons in artificial neural networks}.
\newblock \emph{Distill}.

\bibitem[{Gokaslan and Cohen(2019)}]{gokaslan2019openwebtext}
Aaron Gokaslan and Vanya Cohen. 2019.
\newblock Openwebtext corpus.
\newblock \url{https://skylion007.github.io/OpenWebTextCorpus/}.

\bibitem[{Gurnee et~al.(2023)Gurnee, Nanda, Pauly, Harvey, Troitskii, and Bertsimas}]{gurnee2023finding}
Wes Gurnee, Neel Nanda, Matthew Pauly, Katherine Harvey, Dmitrii Troitskii, and Dimitris Bertsimas. 2023.
\newblock \href {https://openreview.net/forum?id=JYs1R9IMJr} {Finding neurons in a haystack: Case studies with sparse probing}.
\newblock \emph{Transactions on Machine Learning Research}.

\bibitem[{Heap et~al.(2025)Heap, Lawson, Farnik, and Aitchison}]{heap2025sparseautoencodersinterpretrandomly}
Thomas Heap, Tim Lawson, Lucy Farnik, and Laurence Aitchison. 2025.
\newblock \href {https://arxiv.org/abs/2501.17727} {Sparse autoencoders can interpret randomly initialized transformers}.
\newblock \emph{Preprint}, arXiv:2501.17727.

\bibitem[{Huben et~al.(2023)Huben, Cunningham, Smith, Ewart, and Sharkey}]{huben2024sparse}
Robert Huben, Hoagy Cunningham, Logan~Riggs Smith, Aidan Ewart, and Lee Sharkey. 2023.
\newblock \href {https://openreview.net/forum?id=F76bwRSLeK} {Sparse autoencoders find highly interpretable features in language models}.
\newblock In \emph{The Twelfth International Conference on Learning Representations}.

\bibitem[{Kuhn(1955)}]{Kuhn1955TheHM}
Harold~W. Kuhn. 1955.
\newblock \href {https://web.eecs.umich.edu/~pettie/matching/Kuhn-hungarian-assignment.pdf} {The hungarian method for the assignment problem}.
\newblock \emph{Naval Research Logistics (NRL)}, 52.

\bibitem[{Laptev et~al.(2025)Laptev, Balagansky, Aksenov, and Gavrilov}]{laptev2025analyzefeatureflowenhance}
Daniil Laptev, Nikita Balagansky, Yaroslav Aksenov, and Daniil Gavrilov. 2025.
\newblock \href {https://arxiv.org/abs/2502.03032} {Analyze feature flow to enhance interpretation and steering in language models}.
\newblock \emph{Preprint}, arXiv:2502.03032.

\bibitem[{Leask et~al.(2025)Leask, Bussmann, Pearce, Bloom, Tigges, Moubayed, Sharkey, and Nanda}]{leask2025sparse}
Patrick Leask, Bart Bussmann, Michael~T Pearce, Joseph~Isaac Bloom, Curt Tigges, Noura~Al Moubayed, Lee Sharkey, and Neel Nanda. 2025.
\newblock \href {https://openreview.net/forum?id=9ca9eHNrdH} {Sparse autoencoders do not find canonical units of analysis}.
\newblock In \emph{The Thirteenth International Conference on Learning Representations}.

\bibitem[{Li and Eldan(2024)}]{li2024tinystories}
Yuanzhi Li and Ronen Eldan. 2024.
\newblock \href {https://openreview.net/forum?id=yiPtWSrBrN} {Tinystories: How small can language models be and still speak coherent english}.

\bibitem[{Liu et~al.(2024)Liu, Zhan, Lu, Feng, Xue, and Wu}]{liu-etal-2024-good}
Bo~Liu, Li-Ming Zhan, Zexin Lu, Yujie Feng, Lei Xue, and Xiao-Ming Wu. 2024.
\newblock \href {https://aclanthology.org/2024.lrec-main.720/} {How good are {LLM}s at out-of-distribution detection?}
\newblock In \emph{Proceedings of the 2024 Joint International Conference on Computational Linguistics, Language Resources and Evaluation (LREC-COLING 2024)}, pages 8211--8222, Torino, Italia. ELRA and ICCL.

\bibitem[{Longpre et~al.(2023)Longpre, Hou, Vu, Webson, Chung, Tay, Zhou, Le, Zoph, Wei, and Roberts}]{pmlr-v202-longpre23a}
Shayne Longpre, Le~Hou, Tu~Vu, Albert Webson, Hyung~Won Chung, Yi~Tay, Denny Zhou, Quoc~V Le, Barret Zoph, Jason Wei, and Adam Roberts. 2023.
\newblock \href {https://proceedings.mlr.press/v202/longpre23a.html} {The flan collection: Designing data and methods for effective instruction tuning}.
\newblock In \emph{Proceedings of the 40th International Conference on Machine Learning}, volume 202 of \emph{Proceedings of Machine Learning Research}, pages 22631--22648. PMLR.

\bibitem[{Olah et~al.(2020)Olah, Cammarata, Schubert, Goh, Petrov, and Carter}]{olah2020zoom}
Chris Olah, Nick Cammarata, Ludwig Schubert, Gabriel Goh, Michael Petrov, and Shan Carter. 2020.
\newblock \href {https://doi.org/10.23915/distill.00024.001} {Zoom in: An introduction to circuits}.
\newblock \emph{Distill}.
\newblock Https://distill.pub/2020/circuits/zoom-in.

\bibitem[{Olah et~al.(2017)Olah, Mordvintsev, and Schubert}]{olah2017feature}
Chris Olah, Alexander Mordvintsev, and Ludwig Schubert. 2017.
\newblock \href {https://doi.org/10.23915/distill.00007} {Feature visualization}.
\newblock \emph{Distill}.

\bibitem[{Olshausen and Field(1997)}]{olshausen1997sparse}
Bruno~A. Olshausen and David~J. Field. 1997.
\newblock Sparse coding with an overcomplete basis set: A strategy employed by v1?
\newblock \emph{Vision Research}, 37(23):3311--3325.

\bibitem[{Paulo and Belrose(2025)}]{paulo2025sparseautoencoderstraineddata}
Gonçalo Paulo and Nora Belrose. 2025.
\newblock \href {https://arxiv.org/abs/2501.16615} {Sparse autoencoders trained on the same data learn different features}.
\newblock \emph{Preprint}, arXiv:2501.16615.

\bibitem[{Penedo et~al.(2024)Penedo, Kydl{\'\i}{\v{c}}ek, allal, Lozhkov, Mitchell, Raffel, Werra, and Wolf}]{penedo2024the}
Guilherme Penedo, Hynek Kydl{\'\i}{\v{c}}ek, Loubna~Ben allal, Anton Lozhkov, Margaret Mitchell, Colin Raffel, Leandro~Von Werra, and Thomas Wolf. 2024.
\newblock \href {https://openreview.net/forum?id=n6SCkn2QaG} {The fineweb datasets: Decanting the web for the finest text data at scale}.
\newblock In \emph{The Thirty-eight Conference on Neural Information Processing Systems Datasets and Benchmarks Track}.

\bibitem[{Radford et~al.(2019)Radford, Wu, Child, Luan, Amodei, and Sutskever}]{radford2019language}
Alec Radford, Jeffrey Wu, Rewon Child, David Luan, Dario Amodei, and Ilya Sutskever. 2019.
\newblock \href {https://cdn.openai.com/better-language-models/language_models_are_unsupervised_multitask_learners.pdf} {Language models are unsupervised multitask learners}.
\newblock \emph{OpenAI}.
\newblock Accessed: 2024-11-15.

\bibitem[{Rajamanoharan et~al.(2025)Rajamanoharan, Lieberum, Sonnerat, Conmy, Varma, Kramar, and Nanda}]{rajamanoharan2025jumping}
Senthooran Rajamanoharan, Tom Lieberum, Nicolas Sonnerat, Arthur Conmy, Vikrant Varma, Janos Kramar, and Neel Nanda. 2025.
\newblock \href {https://openreview.net/forum?id=mMPaQzgzAN} {Jumping ahead: Improving reconstruction fidelity with jumpre{LU} sparse autoencoders}.

\bibitem[{Schubert et~al.(2021)Schubert, Voss, Cammarata, Goh, and Olah}]{schubert2021high-low}
Ludwig Schubert, Chelsea Voss, Nick Cammarata, Gabriel Goh, and Chris Olah. 2021.
\newblock \href {https://doi.org/10.23915/distill.00024.005} {High-low frequency detectors}.
\newblock \emph{Distill}.

\bibitem[{Sharkey et~al.(2022)Sharkey, Braun, and Millidge}]{sharkey2022sparse}
Lee Sharkey, Dan Braun, and Beren Millidge. 2022.
\newblock \href {https://www.alignmentforum.org/posts/z6QQJbtpkEAX3Aojj/interim-research-report-taking-features-out-of-superposition} {Interim research report: Taking features out of superposition with sparse autoencoders}.
\newblock AI Alignment Forum, posted December 13, 2022.

\bibitem[{Smith et~al.(2025)Smith, Rajamanoharan, Conmy, McDougall, Lieberum, Kramár, Shah, and Nanda}]{smith2025negative}
Lewis Smith, Senthooran Rajamanoharan, Arthur Conmy, Callum McDougall, Tom Lieberum, János Kramár, Rohin Shah, and Neel Nanda. 2025.
\newblock Negative results for saes on downstream tasks and deprioritising sae research.
\newblock \url{https://www.lesswrong.com/posts/4uXCAJNuPKtKBsi28/sae-progress-update-2-draft}.
\newblock DeepMind Mechanistic Interpretability Team Progress Update \#2.

\bibitem[{Socher et~al.(2013)Socher, Perelygin, Wu, Chuang, Manning, Ng, and Potts}]{socher-etal-2013-recursive}
Richard Socher, Alex Perelygin, Jean Wu, Jason Chuang, Christopher~D. Manning, Andrew Ng, and Christopher Potts. 2013.
\newblock \href {https://aclanthology.org/D13-1170/} {Recursive deep models for semantic compositionality over a sentiment treebank}.
\newblock In \emph{Proceedings of the 2013 Conference on Empirical Methods in Natural Language Processing}, pages 1631--1642, Seattle, Washington, USA. Association for Computational Linguistics.

\bibitem[{Stolfo et~al.(2025)Stolfo, Wu, and Sachan}]{stolfo2025antipodal}
Alessandro Stolfo, Ben~Peng Wu, and Mrinmaya Sachan. 2025.
\newblock \href {https://openreview.net/forum?id=Zlx6AlEoB0} {Antipodal pairing and mechanistic signals in dense {SAE} latents}.
\newblock In \emph{ICLR 2025 Workshop on Building Trust in Language Models and Applications}.

\bibitem[{Taori et~al.(2023)Taori, Gulrajani, Zhang, Dubois, Li, Guestrin, Liang, and Hashimoto}]{taori_alpaca_2023}
Rohan Taori, Ishaan Gulrajani, Tianyi Zhang, Yann Dubois, Xuechen Li, Carlos Guestrin, Percy Liang, and Tatsunori~B. Hashimoto. 2023.
\newblock \href {https://crfm.stanford.edu/2023/03/13/alpaca.html} {Alpaca: {A} {Strong}, {Replicable} {Instruction}-{Following} {Model}}.

\bibitem[{Team(2024{\natexlab{a}})}]{gemmateam2024gemma2improvingopen}
Gemma Team. 2024{\natexlab{a}}.
\newblock \href {https://arxiv.org/abs/2408.00118} {Gemma 2: Improving open language models at a practical size}.
\newblock \emph{Preprint}, arXiv:2408.00118.

\bibitem[{Team(2024{\natexlab{b}})}]{grattafiori2024llama3herdmodels}
Llama Team. 2024{\natexlab{b}}.
\newblock \href {https://arxiv.org/abs/2407.21783} {The llama 3 herd of models}.
\newblock \emph{Preprint}, arXiv:2407.21783.

\bibitem[{Templeton et~al.(2024)Templeton, Conerly, Marcus, Lindsey, Bricken, Chen, Pearce, Citro, Ameisen, Jones, Cunningham, Turner, McDougall, MacDiarmid, Freeman, Sumers, Rees, Batson, Jermyn, Carter, Olah, and Henighan}]{templeton2024scaling}
Adly Templeton, Tom Conerly, Jonathan Marcus, Jack Lindsey, Trenton Bricken, Brian Chen, Adam Pearce, Craig Citro, Emmanuel Ameisen, Andy Jones, Hoagy Cunningham, Nicholas~L Turner, Callum McDougall, Monte MacDiarmid, C.~Daniel Freeman, Theodore~R. Sumers, Edward Rees, Joshua Batson, Adam Jermyn, and 3 others. 2024.
\newblock \href {https://transformer-circuits.pub/2024/scaling-monosemanticity/index.html} {Scaling monosemanticity: Extracting interpretable features from claude 3 sonnet}.
\newblock \emph{Transformer Circuits Thread}.

\bibitem[{Vaswani et~al.(2017)Vaswani, Shazeer, Parmar, Uszkoreit, Jones, Gomez, Kaiser, and Polosukhin}]{vaswani2017attention}
Ashish Vaswani, Noam Shazeer, Niki Parmar, Jakob Uszkoreit, Llion Jones, Aidan~N Gomez, \L~ukasz Kaiser, and Illia Polosukhin. 2017.
\newblock \href {https://proceedings.neurips.cc/paper_files/paper/2017/file/3f5ee243547dee91fbd053c1c4a845aa-Paper.pdf} {Attention is all you need}.
\newblock In \emph{Advances in Neural Information Processing Systems}, volume~30. Curran Associates, Inc.

\bibitem[{Wang et~al.(2023)Wang, Ivison, Dasigi, Hessel, Khot, Chandu, Wadden, MacMillan, Smith, Beltagy, and Hajishirzi}]{wang2023how}
Yizhong Wang, Hamish Ivison, Pradeep Dasigi, Jack Hessel, Tushar Khot, Khyathi Chandu, David Wadden, Kelsey MacMillan, Noah~A. Smith, Iz~Beltagy, and Hannaneh Hajishirzi. 2023.
\newblock \href {https://openreview.net/forum?id=w4zZNC4ZaV} {How far can camels go? exploring the state of instruction tuning on open resources}.
\newblock In \emph{Thirty-seventh Conference on Neural Information Processing Systems Datasets and Benchmarks Track}.

\bibitem[{Warstadt et~al.(2019)Warstadt, Singh, and Bowman}]{warstadt-etal-2019-neural}
Alex Warstadt, Amanpreet Singh, and Samuel~R. Bowman. 2019.
\newblock \href {https://doi.org/10.1162/tacl_a_00290} {Neural network acceptability judgments}.
\newblock \emph{Transactions of the Association for Computational Linguistics}, 7:625--641.

\bibitem[{Wei et~al.(2022{\natexlab{a}})Wei, Bosma, Zhao, Guu, Yu, Lester, Du, Dai, and Le}]{wei2022finetuned}
Jason Wei, Maarten Bosma, Vincent Zhao, Kelvin Guu, Adams~Wei Yu, Brian Lester, Nan Du, Andrew~M. Dai, and Quoc~V Le. 2022{\natexlab{a}}.
\newblock \href {https://openreview.net/forum?id=gEZrGCozdqR} {Finetuned language models are zero-shot learners}.
\newblock In \emph{International Conference on Learning Representations}.

\bibitem[{Wei et~al.(2022{\natexlab{b}})Wei, Tay, Bommasani, Raffel, Zoph, Borgeaud, Yogatama, Bosma, Zhou, Metzler, Chi, Hashimoto, Vinyals, Liang, Dean, and Fedus}]{wei2022emergent}
Jason Wei, Yi~Tay, Rishi Bommasani, Colin Raffel, Barret Zoph, Sebastian Borgeaud, Dani Yogatama, Maarten Bosma, Denny Zhou, Donald Metzler, Ed~H. Chi, Tatsunori Hashimoto, Oriol Vinyals, Percy Liang, Jeff Dean, and William Fedus. 2022{\natexlab{b}}.
\newblock \href {https://openreview.net/forum?id=yzkSU5zdwD} {Emergent abilities of large language models}.
\newblock \emph{Transactions on Machine Learning Research}.
\newblock Survey Certification.

\bibitem[{Yang et~al.(2023)Yang, Song, Ren, Lyu, Wang, Zhuo, Liu, Wang, Foster, and Zhang}]{yang-etal-2023-distribution}
Linyi Yang, Yaoxian Song, Xuan Ren, Chenyang Lyu, Yidong Wang, Jingming Zhuo, Lingqiao Liu, Jindong Wang, Jennifer Foster, and Yue Zhang. 2023.
\newblock \href {https://doi.org/10.18653/v1/2023.emnlp-main.276} {Out-of-distribution generalization in natural language processing: Past, present, and future}.
\newblock In \emph{Proceedings of the 2023 Conference on Empirical Methods in Natural Language Processing}, pages 4533--4559, Singapore. Association for Computational Linguistics.

\bibitem[{Zhang et~al.(2015)Zhang, Zhao, and LeCun}]{NIPS2015_250cf8b5}
Xiang Zhang, Junbo Zhao, and Yann LeCun. 2015.
\newblock \href {https://proceedings.neurips.cc/paper_files/paper/2015/file/250cf8b51c773f3f8dc8b4be867a9a02-Paper.pdf} {Character-level convolutional networks for text classification}.
\newblock In \emph{Advances in Neural Information Processing Systems}, volume~28. Curran Associates, Inc.

\end{thebibliography}

\newpage
\onecolumn

\section*{Appendix}
\appendix

The source code for this paper is available at this repository \footnote{\url{https://github.com/seonglae/FaithfulSAE}}.

\section{SAE Training}
\label{appendix:sae-training-hyperparameters}

For the SAE training, the learning rates and TopK values roughly followed the scaling laws proposed by \citet{gao2025scaling}. 100 M tokens were used for all datasets except for LLaMA 8B, where 150 M tokens were used to ensure convergence.
All SAE training was conducted using an NVIDIA RTX 3090ti 24GB. Additionally, to obtain a sufficiently complex feature set when training a single layer, we used the target layer at the 3/4 position except Gemma2 2B model.
For the uncensored instruction dataset, we utilized FLAN\footnote{\url{https://huggingface.co/datasets/Open-Orca/FLAN}}, Open-Instruct \footnote{\url{https://huggingface.co/datasets/xzuyn/open-instruct-uncensored-alpaca}}, and Alpaca dataset \footnote{\url{https://huggingface.co/datasets/aifeifei798/merged_uncensored_alpaca}} in our experiments.

\begin{table*}[h]
\centering
\setlength{\tabcolsep}{6pt}
\begin{tabular}{lcccccrcc}
\toprule
\textbf{Model} & \textbf{Layer} & \textbf{DictSize} & \textbf{TopK} & \textbf{LR} & \textbf{Seed} & \textbf{Dataset} & \textbf{Sequence Length}\\
\midrule
GPT2-small    & 8 & 12288 & 48    & 0.0002   & 42,49 & Faithful-gpt2-small & 128  \\
GPT2-small    & 8 & 12288 & 48    & 0.0002   & 42,49 & Pile-uncopyrighted   & 128  \\
GPT2-small    & 8 & 12288 & 48    & 0.0002   & 42,49 & FineWeb              & 128  \\
GPT2-small    & 8 & 12288 & 48    & 0.0002   & 42,49 & OpenWebText          & 128  \\
GPT2-small    & 8 & 12288 & 48    & 0.0002   & 42,49 & TinyStories          & 128  \\
\midrule
Llama-3.2-1B    & 12 & 14336 & 48    & 0.0002   & 42,49 & Faithful-llama3.2-1b & 512  \\
Llama-3.2-1B    & 12 & 14336 & 48    & 0.0002   & 42,49 & Pile-uncopyrighted   & 512  \\
Llama-3.2-1B    & 12 & 14336 & 48    & 0.0002   & 42,49 & Fineweb              & 512  \\
\midrule
Gemma-2-2b      & 20 & 18432 & 64    & 0.0003   & 42,49 & Faithful-gemma2-2b   & 1024 \\
Gemma-2-2b      & 20 & 18432 & 64    & 0.0003   & 42,49 & Pile-uncopyrighted   & 1024  \\
Gemma-2-2b      & 20 & 18432 & 64    & 0.0003   & 42,49 & Fineweb              & 1024  \\
\midrule
Llama-3.2-3B    & 21 & 18432 & 64    & 0.0001   & 42,49 & Faithful-llama3.2-3b & 512 \\
Llama-3.2-3B    & 21 & 18432 & 64    & 0.0001   & 42,49 & Pile-uncopyrighted   & 512   \\
Llama-3.2-3B    & 21 & 18432 & 64    & 0.0001   & 42,49 & Fineweb              & 512  \\
\midrule
Llama-3.1-8B    & 24 & 16384 & 80    & 6e-05    & 42,49 & Faithful-llama3.1-8b & 512   \\
Llama-3.1-8B    & 24 & 16384 & 80    & 6e-05    & 42,49 & Pile-uncopyrighted   & 512  \\
Llama-3.1-8B    & 24 & 16384 & 80    & 6e-05    & 42,49 & Fineweb              & 512  \\
\midrule
Pythia-1.4B    & 18 & 14336 & 48    & 0.0002    & 42,49 & Faithful-pythia-1.4b               & 512  \\
Pythia-1.4B    & 18 & 14336 & 48    & 0.0002    & 42,49 & Faithful-pythia-2.8b               & 512  \\
Pythia-1.4B    & 18 & 14336 & 48    & 0.0002    & 42,49 & Open-Instruct             & 512  \\
Pythia-1.4B    & 18 & 14336 & 48    & 0.0002    & 42,49 & Alpaca-Instruction             & 512  \\
Pythia-1.4B    & 18 & 14336 & 48    & 0.0002    & 42,49 & FLAN             & 512  \\
\midrule
Pythia-2.8B    & 24 & 15360 & 64    & 0.0001    & 42,49 & Faithful-pythia-1.4b           & 512  \\
Pythia-2.8B    & 24 & 15360 & 64    & 0.0001    & 42,49 & Faithful-pythia-2.8b            & 512  \\
Pythia-2.8B    & 24 & 15360 & 64    & 0.0001    & 42,49 & Open-Instruct         & 512  \\
Pythia-2.8B    & 24 & 15360 & 64    & 0.0001    & 42,49 & Alpaca-instruction          & 512  \\
Pythia-2.8B    & 24 & 15360 & 64    & 0.0001    & 42,49 & FLAN          & 512  \\
\bottomrule
\end{tabular}
\caption{SAE training hyperparameters for each model and dataset. The configuration includes the model name, layer index, dictionary size, top-$k$ sparsity, learning rate, random seed, training dataset, and sequence/token dimensions. (a) and (b) are shorthand tags used for table compactness.}
\label{tab:file-configurations}
\end{table*}

\newpage
\section{Faithful SAEs}
\label{sec:faithful-saes}

The figures below show how each SAE trained on different datasets generalizes its reconstruction capability on other datasets, demonstrating its faithfulness. They compare the Explained Variance, L2 loss, and CE difference across datasets when the LLM’s hidden state is replaced by the SAE’s reconstructed activation trained on a specific dataset. The X-axis represents the evaluation dataset, and the Y-axis indicates the SAE’s training dataset. All results are based on SAE models trained with seed 42. The trained SAEs are available in the following collection \footnote{\url{https://huggingface.co/collections/seonglae/faithful-saes-67f3b25ff21a185017879b33}}.

\begin{figure}[h]
  \centering
  \includegraphics[width=0.9\linewidth]{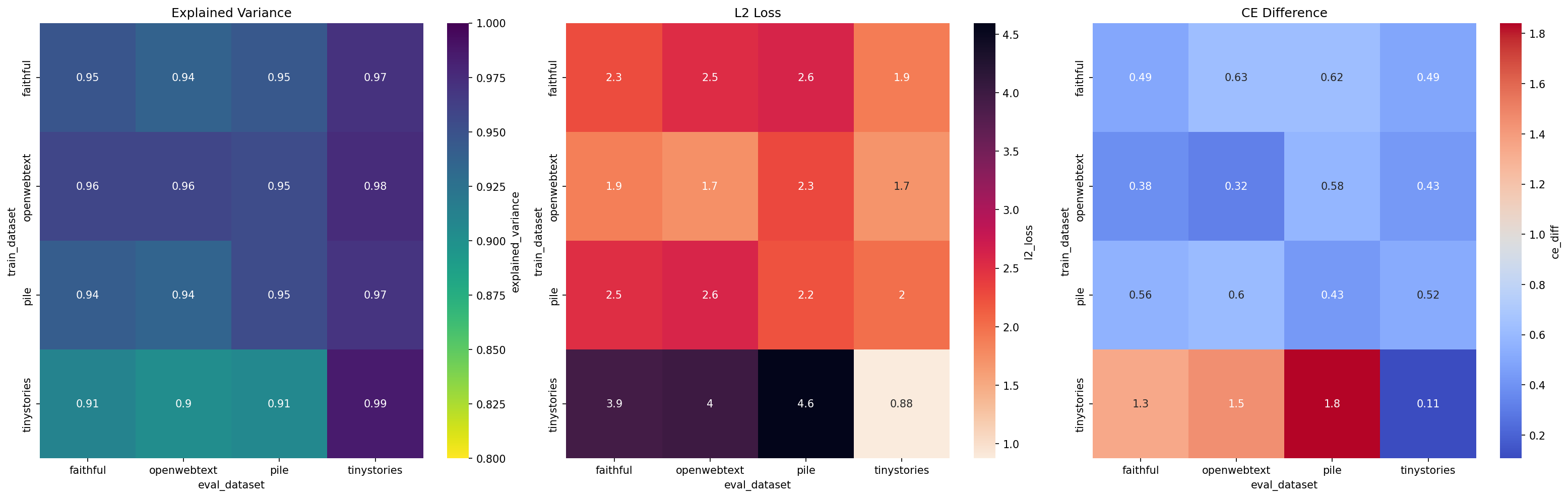}
  \caption{Faithful SAE representation for GPT-2. This figure visualizes the SAE model's ability to reconstruct GPT-2's hidden state.}
  \label{fig:sae-gpt2}
\end{figure}

\begin{figure}[h]
  \centering
  \includegraphics[width=0.9\linewidth]{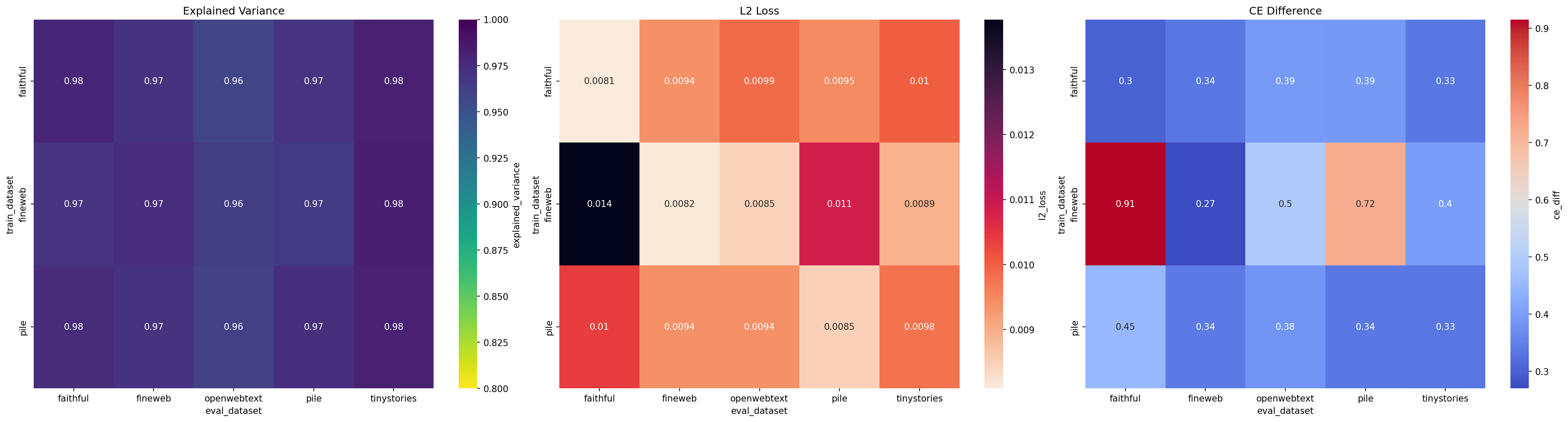}
  \caption{Faithful SAE representation for LLaMA 1B. This figure demonstrates the SAE's performance in reconstructing the hidden state of LLaMA 1B.}
  \label{fig:sae-llama1}
\end{figure}

\begin{figure}[h]
  \centering
  \includegraphics[width=0.9\linewidth]{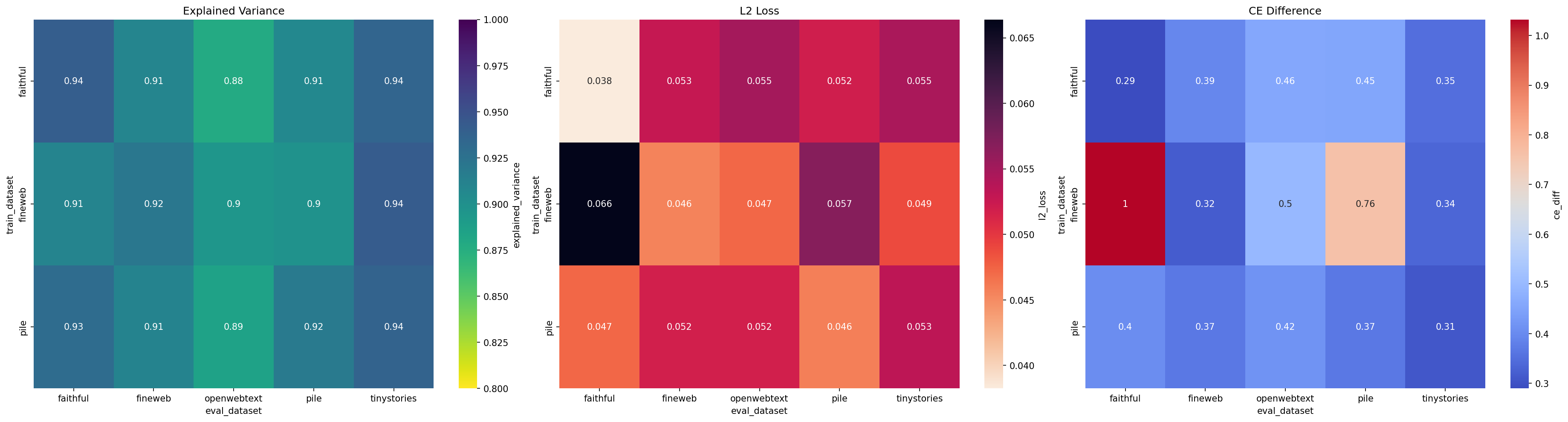}
  \caption{Faithful SAE representation for LLaMA 3B. This figure highlights the SAE's reconstruction quality for the LLaMA 3B model's hidden state.}
  \label{fig:sae-llama3}
\end{figure}

\begin{figure}[H]
  \centering
  \includegraphics[width=0.9\linewidth]{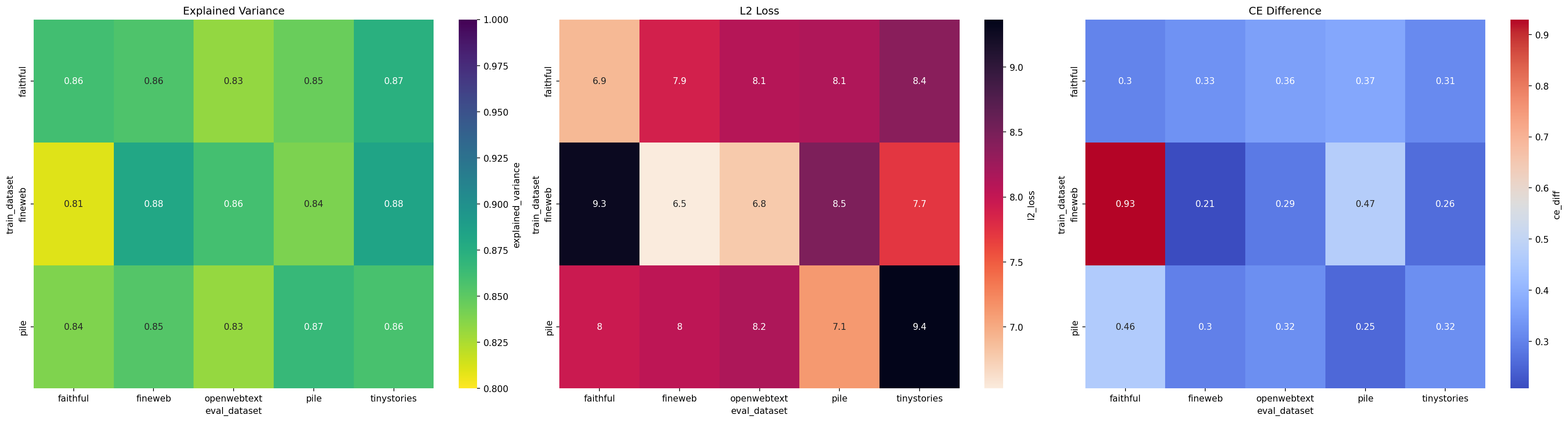}
  \caption{Faithful SAE representation for Gemma 2B. This figure shows the SAE's reconstruction of the Gemma 2B hidden state and its faithfulness across datasets. }
  \label{fig:sae-gemma2}
\end{figure}

\section{Faithful Dataset}
\label{subsec:faithful-dataset}

The figures below compare the model's BOS token's next token distribution and the empirical frequency distribution of the first token from our generated Faithful dataset. The left two figures represent the model's distribution, and the right two figures represent the dataset's token frequency distribution. The upper two figures show only the top 10 tokens, which show almost identical shapes to the original model. However, the bottom two graphs show that the frequency distribution does not cover the whole token distribution, as the probability decreases exponentially for the first generation. By comparing the coverage and token statistics, we verified that the Faithful dataset reflects the original model's capability well. Additionally, the Pythia 6.9B model was used solely to generate dataset and to verify that the first token distribution matches the model's BOS token and was not used for training.  The Faithful datasets are available in the following collection \footnote{\url{https://huggingface.co/collections/seonglae/faithful-dataset-67f3b21ff8fca56b87e5370f}}.

\begin{figure}[H]
  \centering
  \includegraphics[width=0.8\linewidth]{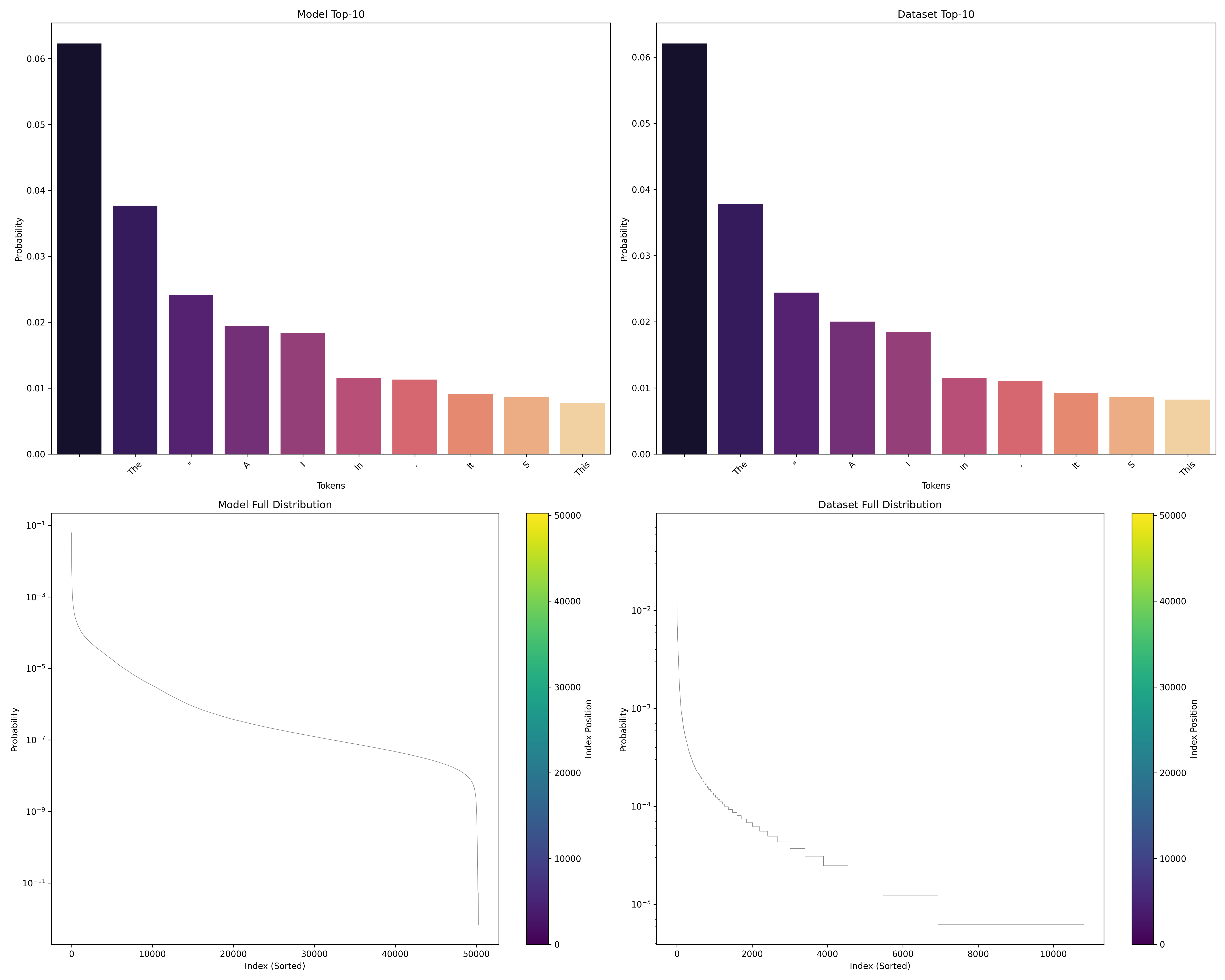}
  \caption{This figure compares the token distribution of the generated dataset for GPT-2 with the model's expected token distribution.}
\end{figure}

\begin{figure}[H]
  \centering
  \includegraphics[width=0.8\linewidth]{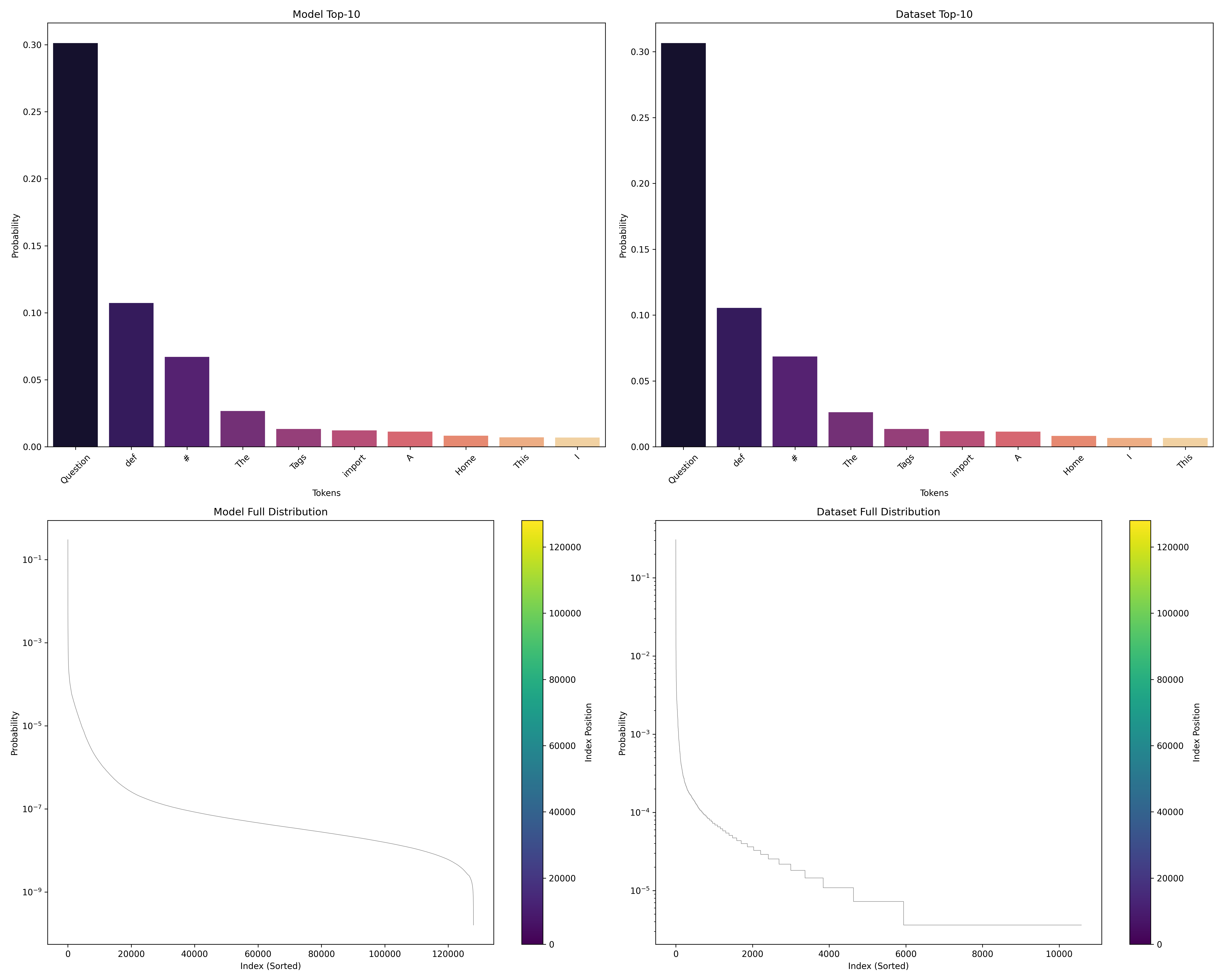}
  \caption{This figure compares the token distribution of the generated dataset for LLaMA 1B with the model's original token distribution.}
\end{figure}

\begin{figure}[h]
  \centering
  \includegraphics[width=0.8\linewidth]{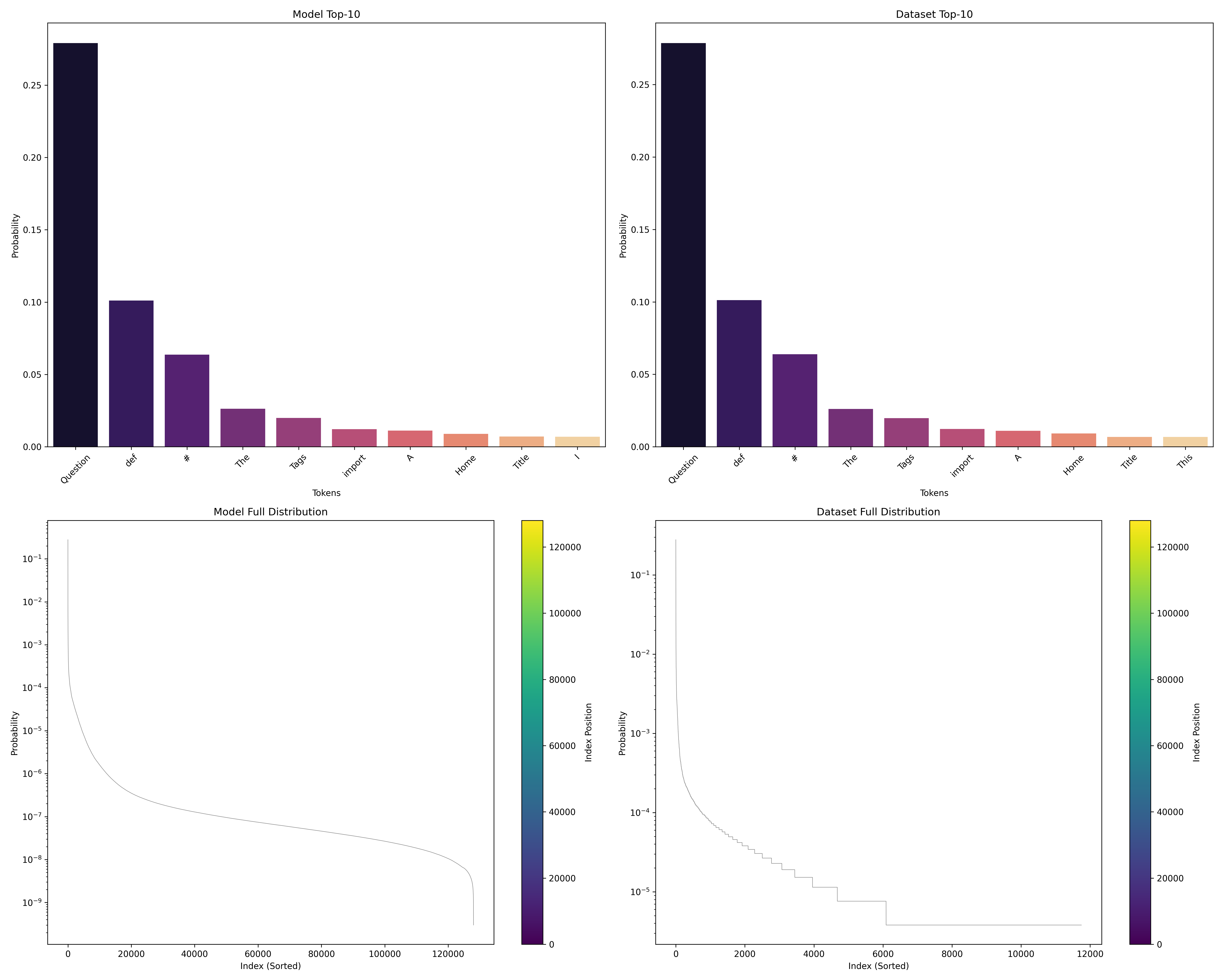}
  \caption{This comparison shows the token distribution of LLaMA 3B's generated dataset versus the model's distribution.}
\end{figure}

\begin{figure}[H]
  \centering
  \includegraphics[width=0.8\linewidth]{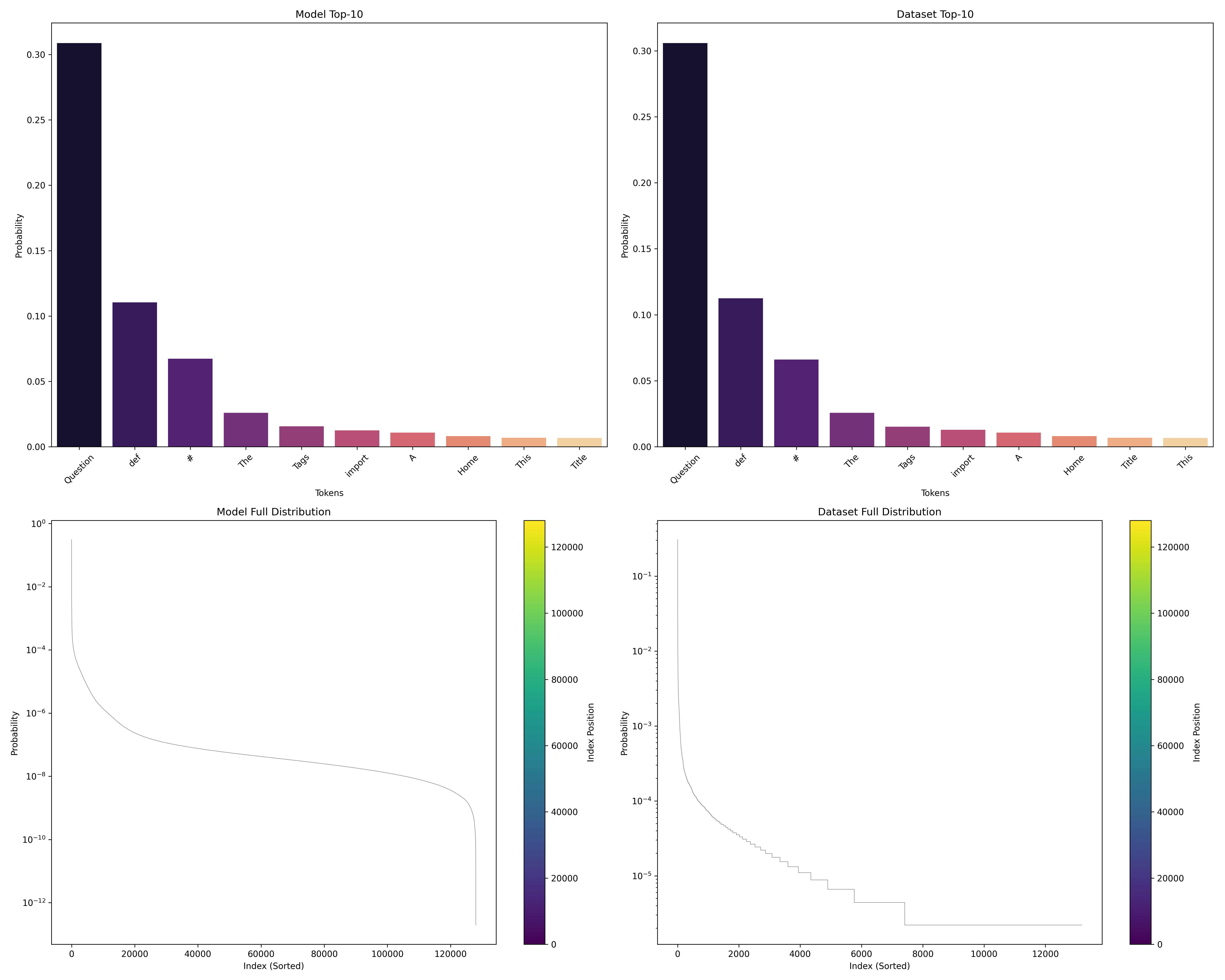}
  \caption{This figure visualizes how well the generated dataset represents LLaMA 8B's token distribution.}
\end{figure}

\begin{figure}[h]
  \centering
  \includegraphics[width=0.8\linewidth]{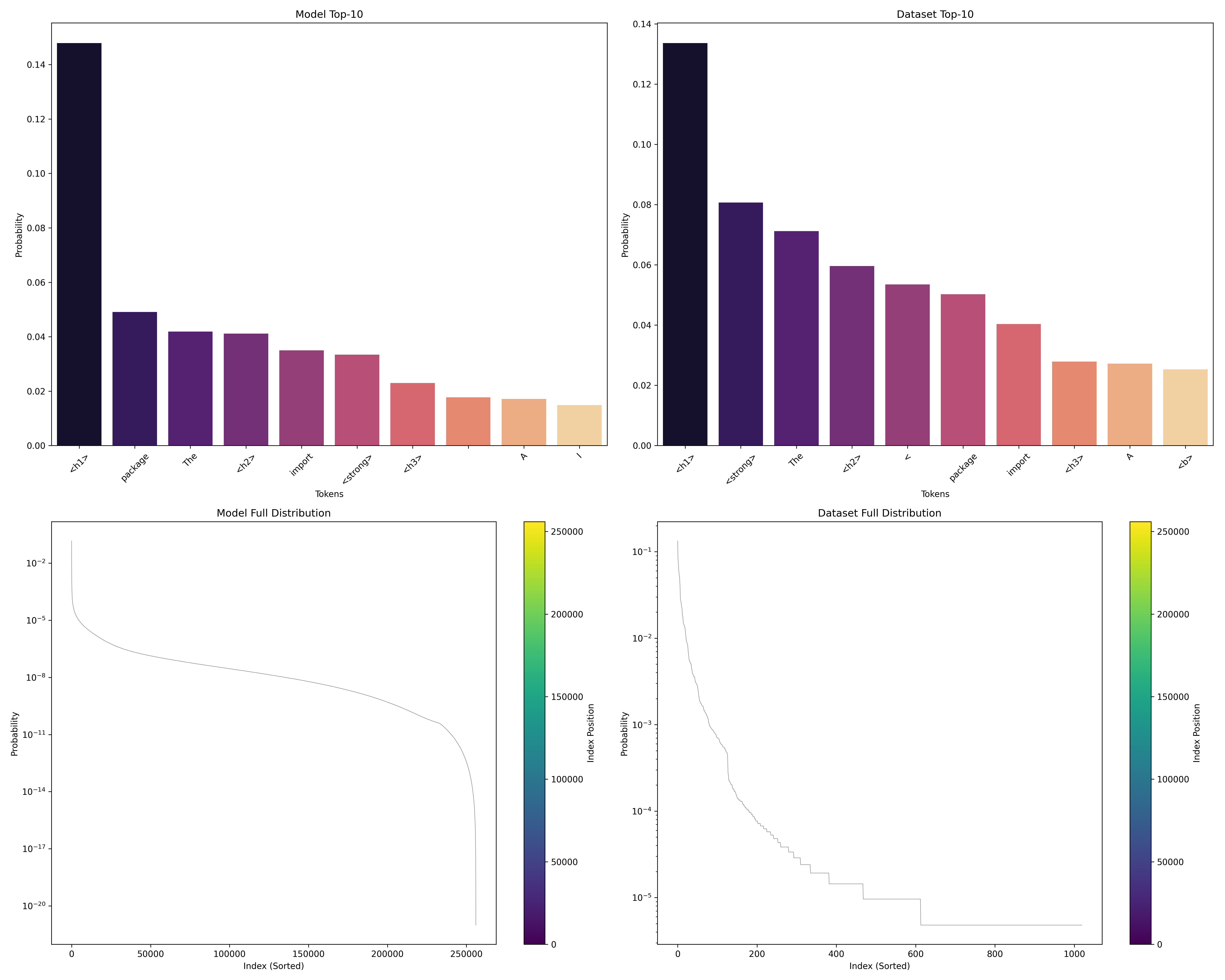}
  \caption{This visualization compares the generated token distribution with the original model for Gemma 2B.}
\end{figure}

\begin{figure}[H]
  \centering
  \includegraphics[width=0.8\linewidth]{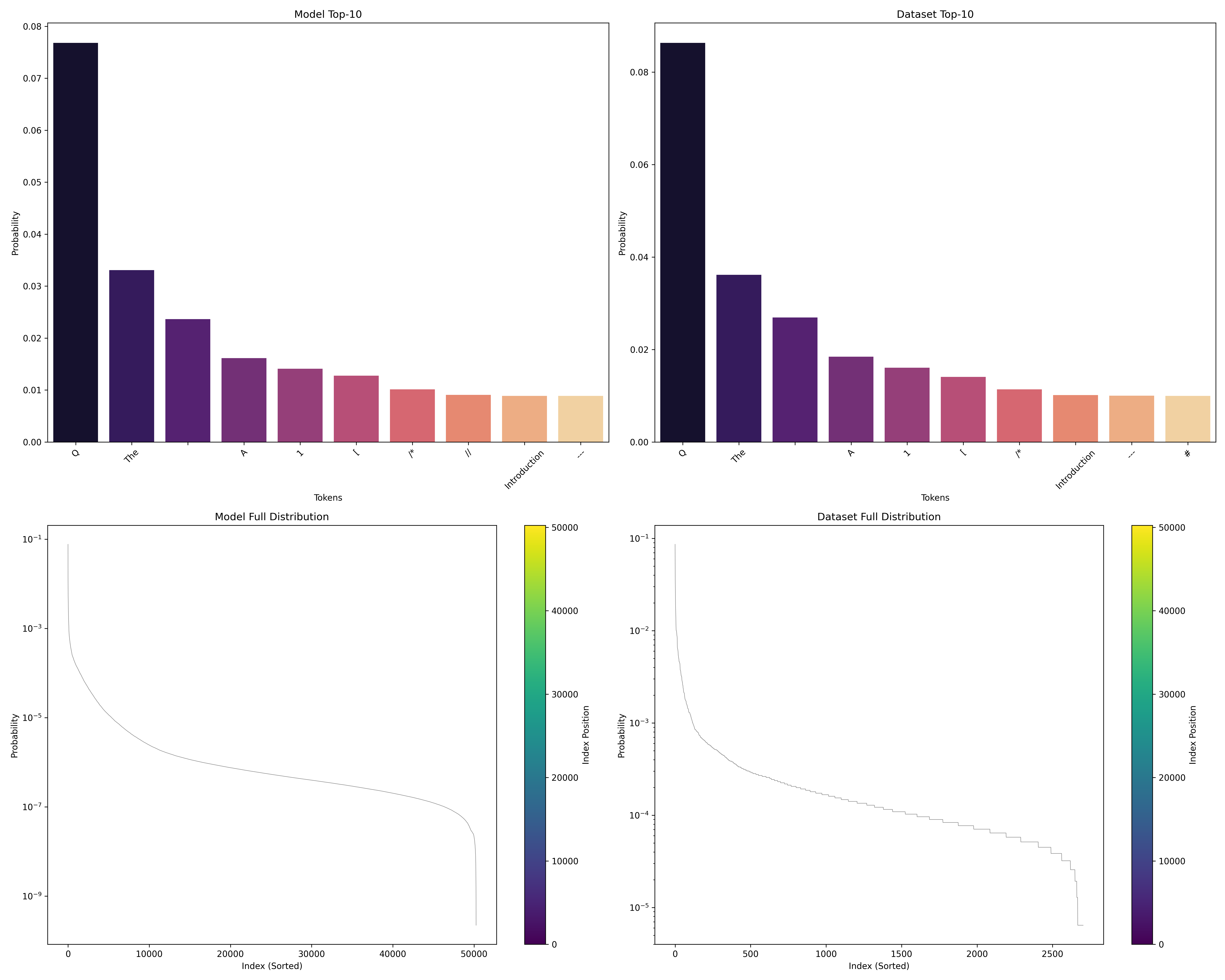}
  \caption{This figure shows the token distribution for the generated Pythia 1.4B dataset, comparing it to the model's distribution.}
\end{figure}

\begin{figure}[H]
  \centering
  \includegraphics[width=0.8\linewidth]{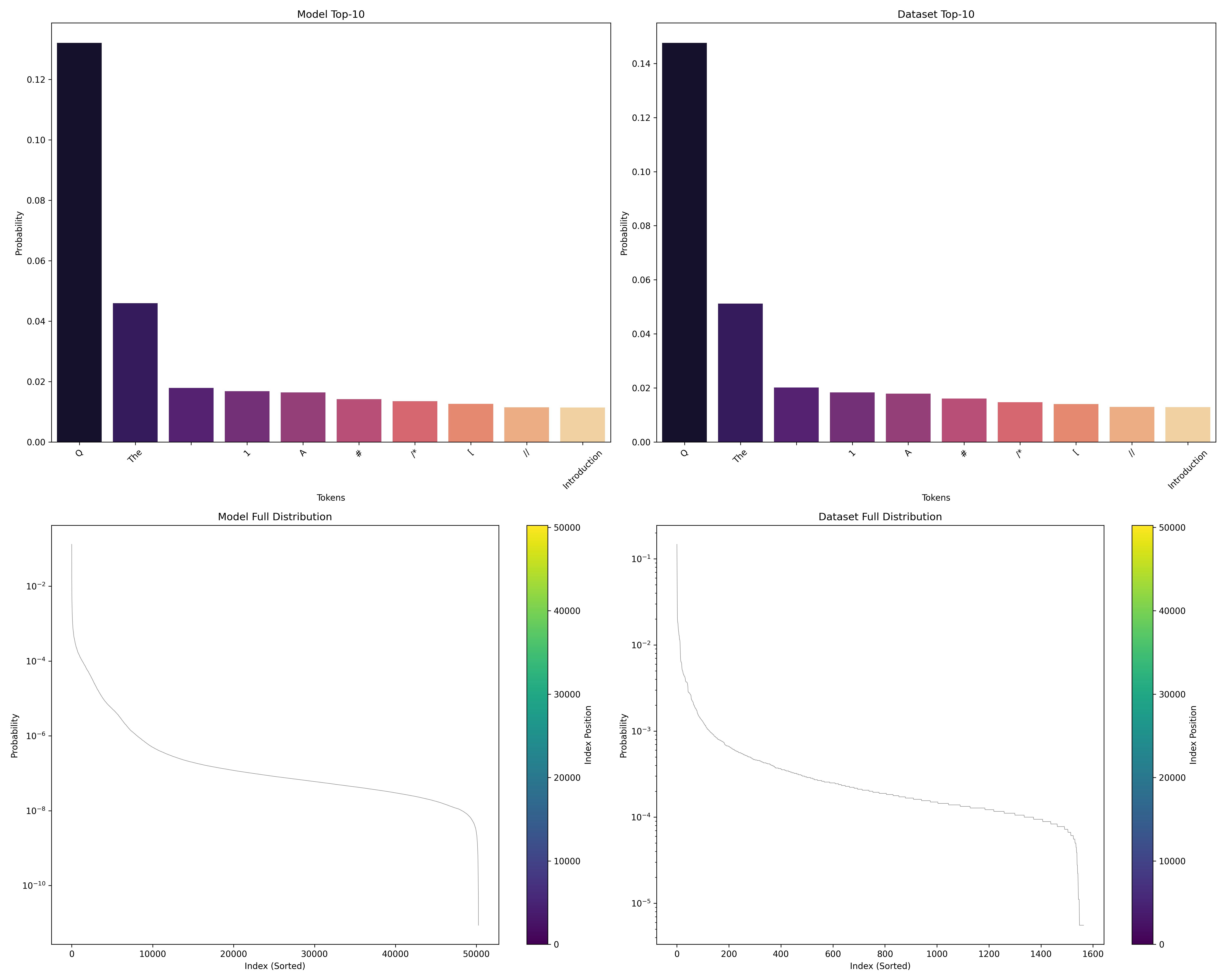}
  \caption{This figure shows the token distribution for the generated Pythia 2.8B dataset, comparing it to the model's distribution.}
\end{figure}

\begin{figure}[H]
  \centering
  \includegraphics[width=0.8\linewidth]{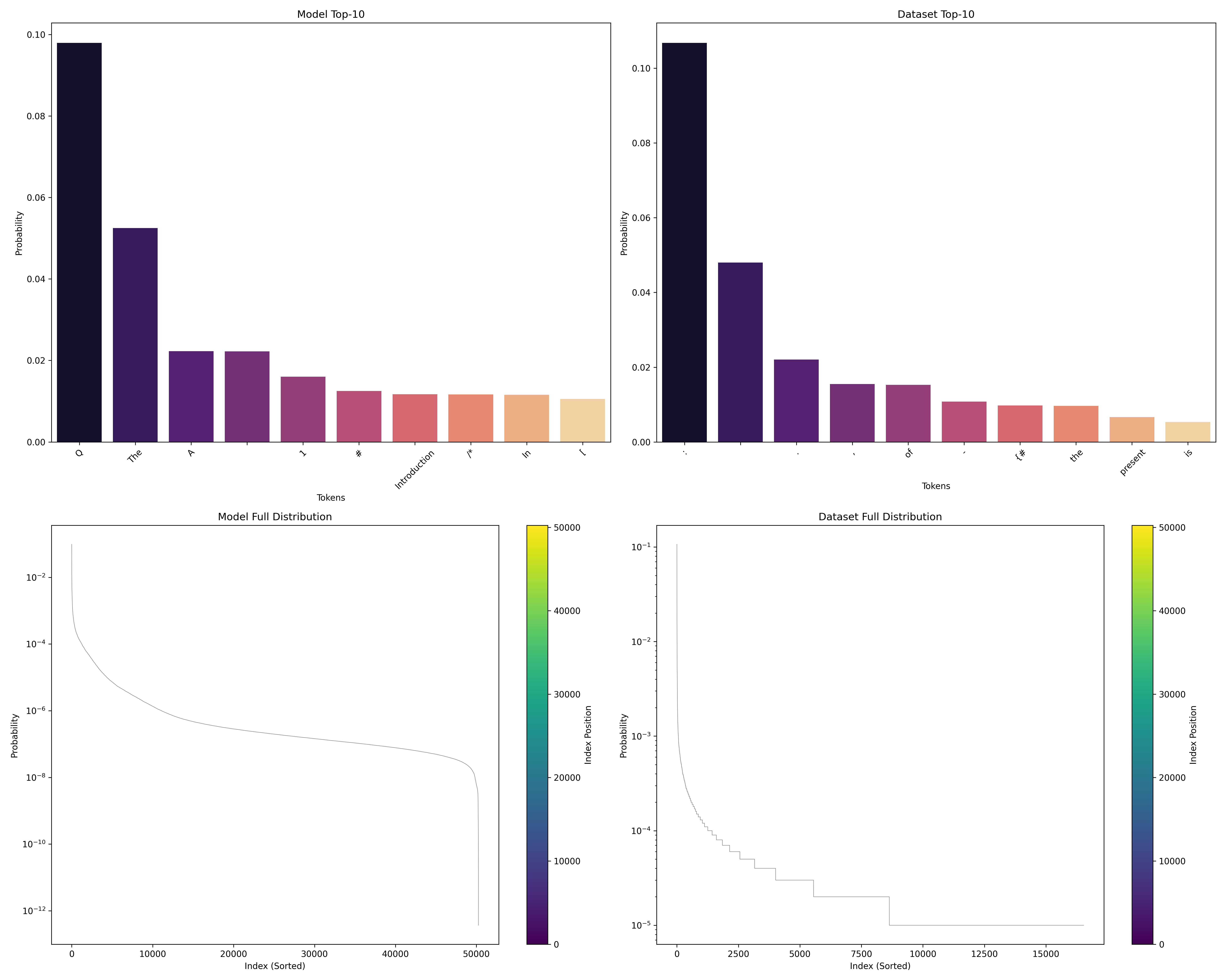}
  \caption{This figure shows the token distribution for the generated Pythia 6.9B dataset, comparing it to the model's distribution.}
\end{figure}

\subsection{SAE Probing}

\begin{table}[H]
\centering
\begin{tabular}{lccc|ccc|ccc}
\toprule
\multirow{2}{*}{Model} & \multicolumn{3}{c|}{SST-2} & \multicolumn{3}{c|}{CoLA} & \multicolumn{3}{c}{Yelp} \\
                       & Faithful & Fineweb & Pile & Faithful & Fineweb & Pile & Faithful & Fineweb & Pile \\
\midrule
GPT2-small   & \textbf{0.7746} & 0.7723 & 0.7500 & \textbf{0.7076} & 0.6989 & 0.6912 & \textbf{0.6532} & 0.6502 & 0.6444 \\
Pythia 1.4B  & \textbf{0.8451} & 0.8354 & 0.8314 & \textbf{0.7281} & 0.7253 & 0.7262 & 0.9341 & \textbf{0.9399} & 0.9289 \\
Gemma 2B     & 0.7729 & \textbf{0.8394} & 0.8085 & \textbf{0.7478} & 0.7291 & 0.7430 & \textbf{0.9536} & 0.9495 & 0.9440 \\
Pythia 2.8B  & 0.8050 & 0.8256 & \textbf{0.8365} & \textbf{0.6985} & 0.6371 & 0.6783 & 0.9392 & 0.9428 & \textbf{0.9442} \\
LLaMA 1B     & 0.8342 & \textbf{0.8491} & 0.8428 & \textbf{0.7469} & 0.7411 & 0.7411 & 0.9431 & \textbf{0.9437} & 0.9429 \\
LLaMA 3B     & \textbf{0.8532} & 0.8423 & 0.8497 & \textbf{0.6889} & 0.6826 & 0.6888 & \textbf{0.9547} & 0.9544 & 0.9525 \\
\bottomrule
\end{tabular}
\caption{Reconstruction accuracy of SAE probing across 3 datasets and 6 model architectures. FaithfulSAE compared against SAEs trained on web-based datasets (Fineweb, Pile).}
\label{tab:sae-probing-table}
\end{table}

\subsection{Fake Feature}

\begin{table}[H]
\centering
\begin{tabular}{lccccccc}
\toprule
\textbf{Dataset} & \textbf{GPT2} & \textbf{Pythia 1.4B} & \textbf{Gemma 2B} & \textbf{Pythia 2.8B} & \textbf{LLaMA 1B} & \textbf{LLaMA 3B} & \textbf{LLaMA 8B} \\
\midrule
Faithful & \textbf{0.1139} & 0.3871 & \textbf{0.5425} & 0.4655 & \textbf{0.0314} & \textbf{0.1899} & \textbf{0.4150} \\
Pile     & 0.1180 & 0.3871 & 0.5669 & 0.4460 & 0.0446 & 0.2930 & 0.5341 \\
Fineweb  & 0.1587 & \textbf{0.3802} & 0.5995 & \textbf{0.4362} & 0.0600 & 0.2713 & 0.5493 \\
\bottomrule
\end{tabular}
\caption{Average fake feature ratio (\%) across training datasets and model architectures.}
\label{tab:fake-feature}
\end{table}

\end{document}